\documentclass[sn-basic]{sn-jnl}

\usepackage{graphicx}
\usepackage{multirow}
\usepackage{amsmath,amssymb,amsfonts}
\usepackage{amsthm}
\usepackage{mathrsfs}
\usepackage[title]{appendix}
\usepackage{xcolor}
\usepackage{textcomp}
\usepackage{manyfoot}
\usepackage{booktabs}
\usepackage{algorithm}
\usepackage{algorithmicx}
\usepackage{algpseudocode}
\usepackage{listings}
\usepackage{array} 
\usepackage{makecell} 
\usepackage{longtable, booktabs, lscape} 
\usepackage{caption}
\usepackage{ragged2e}
\usepackage{footmisc}
\usepackage{lipsum}

\begin{document}

\title[Article Title]{A Generic Review of Integrating Artificial Intelligence in Cognitive Behavioral Therapy}

\author[1]{\fnm{Meng} \sur{Jiang}}
\author[1]{\fnm{Qing} \sur{Zhao}}
\author[1]{\fnm{Jianqiang} \sur{Li}}
\author[2]{\fnm{Fan} \sur{Wang}}
\author[2]{\fnm{Tianyu} \sur{He}}
\author[2]{\fnm{Xinyan} \sur{Cheng}}
\author[2]{\fnm{Bing Xiang} \sur{Yang}}
\author[3]{\fnm{Grace W.K.} \sur{Ho}}
\author*[4]{\fnm{Guanghui} \sur{Fu}}\email{guanghui.fu@icm-institute.org}

\affil[1]{\orgdiv{School of Software Engineering}, \orgname{Beijing University of Technology}, \orgaddress{\city{Beijing}, \country{China}}}

\affil[2]{\orgdiv{School of Nursing}, \orgname{Wuhan University}, \orgaddress{\city{Wuhan}, \country{China}}}

\affil[3]{\orgdiv{School of Nursing}, \orgname{Joint Research Centre for Primary Health Care, The Hong Kong Polytechnic University}, \orgaddress{\city{Hong Kong}, \country{China}}}

\affil[4]{\orgdiv{Sorbonne Université}, \orgname{ICM, CNRS, Inria, Inserm, AP-HP, Hôpital de la Pitié-Salpêtrière}, \orgaddress{\city{Paris}, \country{France}}}

\abstract{Cognitive Behavioral Therapy (CBT) is a well-established intervention for mitigating psychological issues by modifying maladaptive cognitive and behavioral patterns. However, delivery of CBT is often constrained by resource limitations and barriers to access. Advancements in artificial intelligence (AI) have provided technical support for the digital transformation of CBT. Particularly, the emergence of pre-training models (PTMs) and large language models (LLMs) holds immense potential to support, augment, optimize and automate CBT delivery. This paper reviews the literature on integrating AI into CBT interventions. 
We begin with an overview of CBT. Then, we introduce the integration of AI into CBT across various stages: pre-treatment, therapeutic process, and post-treatment. Next, we summarized the datasets relevant to some CBT-related tasks. Finally, we discuss the benefits and current limitations of applying AI to CBT. We suggest key areas for future research, highlighting the need for further exploration and validation of the long-term efficacy and clinical utility of AI-enhanced CBT. The transformative potential of AI in reshaping the practice of CBT heralds a new era of more accessible, efficient, and personalized mental health interventions.
}


\maketitle

\section{Introduction}\label{sec:intro}

Mental health issues have become increasingly prevalent, especially in the post-pandemic era~\cite{penninx2022covid}. Psychological distress brought on by COVID-19 has added to the existing mental health crisis, leading to higher rates of depression, anxiety, and suicidal ideation. Consequently, there is an urgent need for effective and accessible mental health interventions.
Cognitive Behavioral Therapy (CBT) is widely recognized as one of the most important psychological interventions for addressing a range of mental health issues~\cite{foreman2016cognitive, david2018cognitive}. As the gold standard for treating depression and anxiety, CBT has been integrated into healthcare systems worldwide. Its primary goal is to identify and restructure patients' maladaptive cognitive frameworks to help them develop coping skills, address behavioral issues, and alleviate symptoms. However, the delivery of CBT, as it is traditionally conducted individually and face-to-face by a therapist, faces barriers in real life, including social stigma and limited access to qualified therapists, particularly in underserved areas. Only 27\% of those receiving psychological therapy actually access standardized care~\cite{bandelow2017treatment}. To address these challenges, technological advancements have led to the development of Computer-Based Cognitive Behavioral Therapy (CCBT) and Internet-Based Cognitive Behavioral Therapy (ICBT)~\cite{webb2017internet}. These variations use computer software and the Internet as a medium to deliver CBT interventions based on established theories and techniques. Several established online CBT platforms, such as MoodGYM~\cite{MoodGym} and 30-Day Self-Service Psychological Expert~\cite{30DaySelfService}, have emerged. These online platforms have partially alleviated some of the issues related to the shortage of mental health professionals and poor accessibility~\cite{christ2020internet}. However, they still face challenges, such as limited interactivity and high dropout rates. In addition, they offer interventions that address only general cognitive issues rather than individual specific needs. Given this challenge, there is an urgent need for more flexible and adaptive forms of CBT to better meet the diverse needs of different patient populations.

In recent years, rapid advancements in artificial intelligence (AI) technology have led various industries to explore its innovative applications. AI's exceptional capabilities in data analysis, pattern recognition, and automation offer vast potential in the delivery of mental health services, providing new tools and methods for assessment, diagnosis, and treatment of mental health conditions~\cite{lee2021artificial, malgaroli2023natural, 2023Using}. 
For instance, deep learning (DL) models can automate mental health assessments and diagnostic processes, alleviating the workload of professionals while enhancing the accuracy and efficiency of diagnoses~\cite{vuyyuru2023transformer}. Furthermore, AI can also aid in detecting and understanding patients' unique emotional expressions, thereby offering recommendations for personalized psychological interventions and support~\cite{assunccao2022overview}. Given the shortage of skilled mental health professionals, innovative AI approaches have been developed to guide peer-to-peer mental health support~\cite{sharma2023human}. Additionally, AI-based psychological counseling support systems, such as those utilizing large language models (LLMs), have also been developed to assist junior counselors in providing online psychological support~\cite{fu2023enhancing}. These studies demonstrate that AI not only expands the boundaries of traditional mental health services, but also brings new opportunities for improving their accessibility and utility globally~\cite{graham2019artificial, demszky2023using}.
In the realm of CBT, the integration of AI has led to revolutionary advancements, particularly with the rise of LLMs. Notable research in this area includes the development of CBT-specific prompts and tailored datasets, leading to models like CBT-LLM, which are specifically designed for CBT delivery~\cite{na2024cbt}. Other efforts have resulted in conversational counseling agents based on LLMs, such as utilizing GPT-2 in CBT to generate human-like textual narratives to provide psychological support~\cite{rajagopal2021novel}, providing psychological counseling using CBT techniques~\cite{lee2024cocoa}, and delivering dialogue modules focused on Socratic questioning using LLMs like OsakaED and GPT-4~\cite{izumi2024response}. These advancements have made cognitive behavioral interventions more personalized and precise, better addressing the needs of diverse patients. 

Given the abundance of research emerging in this field, several review articles have summarized innovative developments in CBT. However, the exsiting reviews focused mainly on CBT applications~\cite{huguet2016systematic, denecke2022implementation} and the development of ChatGPT in redesigning CBT for different age groups and genders~\cite{chandra2023chatgpt}. There is a lack of comprehensive review summarizing the application of AI's role across various stages of CBT delivery. To address this, we provide a detailed literature review of AI's role in enhancing CBT in this paper. 

\section{Background} \label{sec:background} 

Cognitive behavioral therapy (CBT) stands as a structured, time-limited psychological treatment that focuses on the interplay between cognitive processes, emotional responses, and behavioral patterns. This therapeutic approach operates on the premise that thoughts, emotions, and behaviors are interconnected, and aims to enhance psychological well-being by addressing maladaptive cognitive patterns and behaviors~\cite{beck2011cognitive}. 
The theoretical foundation of CBT encompasses both cognitive and behavioral aspects. Regarding the origins of CBT, there are varying views in the academic community. However, it is generally accepted that the popularization of the ``cognition'' began with Ellis’s Rational Emotive Therapy (RET) in the 1950s, followed by Beck’s Cognitive Therapy (CT) in the 1960s. Since then, CBT has continuously integrated various behavioral therapies and theories, evolving into the widely used psychological treatment seen today.

CBT interventions typically target three domains: cognition, behavior, and emotion~\cite{mcginn2001allows}. 
In the cognitive domain, CBT focuses on individuals' cognitive processes, including thoughts, beliefs, and interpretations. This involves aiding clients in identifying and understanding negative or distorted thought patterns, such as negative automatic thoughts and cognitive distortions (CD) (e.g., jumping to conclusions, all-or-nothing thinking and mental filtering). Techniques like cognitive restructuring (CR) are used to adjust these patterns, fostering a more objective and positive perspective toward oneself and the world. In the behavioral domain, CBT addresses individuals' maladaptive behavioral patterns and habits. Therapists collaborate with clients to explore their unhealthy behaviors, such as social withdrawal, avoidance, and substance or alcohol misuse. Techniques such as behavioral experiments, graded exposure, and behavioral activation (BA) are then employed. Further, CBT acknowledges the interplay between the body and the mind. Physiological responses such as tension, fear, and anxiety can impact cognition and behavior. Thus, CBT also focuses on regulating physiological responses through techniques such as deep breathing, progressive muscle relaxation, and physical activity to enhance emotional regulation and psychological well-being. For the emotional domain, CBT targets clients’ emotional awareness and ways to manage them by aiding clients to recognize negative emotions as they arise, accurately identify them, and utilize appropriate cognitive and behavioral strategies to support better emotional well-being. The three domains are intertwined, and CBT typically tailors a range of techniques and strategies based on the individual's specific circumstances and needs, aiding in problem-solving, improving mental health, and building resilience~\cite{dobson2021handbook}.

CBT has been employed to address a variety of mental health issues, including anxiety disorder~\cite{olatunji2010efficacy}, depression~\cite{tymofiyeva2019application}, schizophrenia~\cite{turkington2004cognitive}, attention deficit and hyperactivity disorder~\cite{pan2019comparison}, insomnia~\cite{benard2018q}, eating disorders~\cite{linardon2017efficacy}, bipolar disorder~\cite{driessen2010cognitive}, substance use disorders~\cite{mchugh2010cognitive}, and obsessive-compulsive disorder~\cite{moody2017mechanisms}. Beyond mental health, CBT has been explored as a tool for managing chronic health conditions like low back pain~\cite{piette2016patient, heapy2017interactive}, asthma~\cite{parry2012cognitive}, and tinnitus~\cite{parry2012cognitive}. Prior to initiating CBT, therapists typically conduct assessments to determine symptom severity and treatment goals. These assessments may involve dialogue between therapist and client or self-administered tools like the Beck Depression Inventory (BDI-II) or the Beck Anxiety Inventory (BAI)~\cite{bech1996beck, beck1988inventory, foa1993reliability}. Based on this assessment, therapists work with the clients to establish treatment goals and offer personalized cognitive and behavioral strategies tailored to their needs.

\section{Artificial intelligence in CBT} \label{sec:ai_cbt}
\begin{figure}[!hbtp]
\centering
\includegraphics[width=1.0\linewidth]{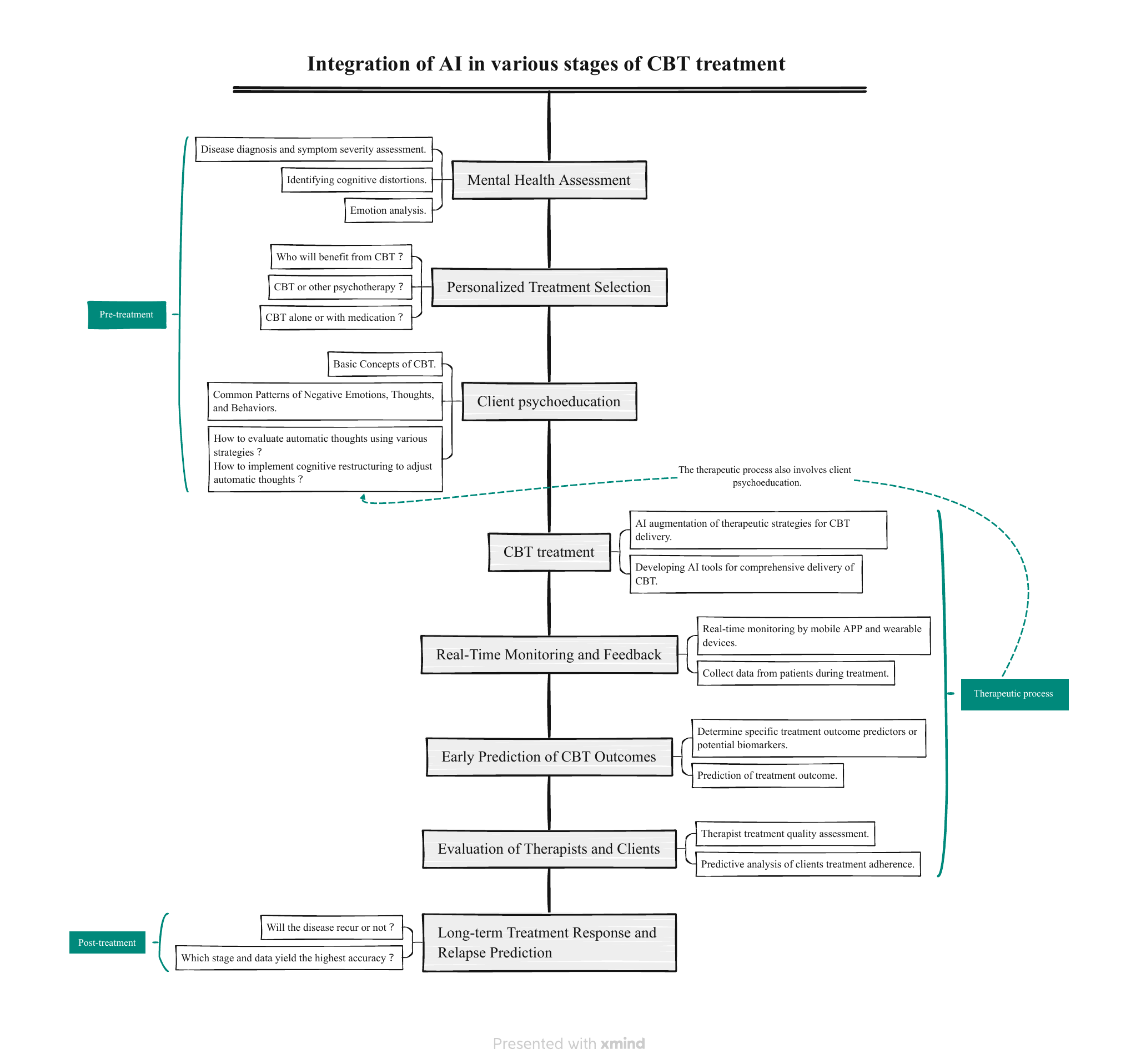}
\caption{The role of AI in various stages of CBT treatment.}
\label{fig:ai_cbt_Stages}
\end{figure}

Integrating AI into CBT presents a myriad of potential applications, promising to enhance the effectiveness and utility of therapy. 
To provide a summary of the extant literature on the integration of AI into CBT across the delivery process, we conducted a literature review by performing searches on the ArXiv and Google Scholar database without date restrictions. Search terms considered in this review were selected based on two elements, CBT (e.g., cognitive behavioral therapy, cognitive distortions, cognitive restructuring, and exposure therapy) and technology (e.g., artificial intelligence, machine learning, deep learning, natural language processing, large language model, chatbot, and virtual reality). Articles that enhance, support, or implement CBT through AI technology will be selected. Results were categorized based on the stage at which AI was integrated in the CBT delivery process and were synthesized.
In this section, we present the synthesized findings describing the role of AI in different stages of the CBT treatment process, as illustrated in Figure~\ref{fig:ai_cbt_Stages}. 

\textbf{To be clear, this article only covers the stages of CBT that current AI technology has addressed. Due to the complexity of CBT treatment, some details and specific treatment stages have not yet been covered by AI, and thus are not discussed here. Future research and technological advancements may further expand AI's role in CBT delivery, encompassing more stages of the treatment process.}

\subsection{Integration of AI in the Pre-Treatment Stage}

A comprehensive assessment is a foundational step in all psychological treatments, including CBT. This initial assessment involves gathering information on the client’s history, current issues, and therapeutic goals through structured or semi-structured interviews, questionnaires, and standardized scales. In CBT, particular emphasis is placed on assessing the client's cognitive and behavioral patterns, as well as their emotional responses. The insights from this detailed assessment guides therapists in crafting tailored CBT treatment plans that effectively address the client's concerns, thereby maximizing therapeutic outcomes.
Traditional CBT assessment methods, while thorough, often rely heavily on manual processes, which can be subjective and time-consuming. The reliance on the therapist’s clinical skills and professional acumen can may also limit diagnostic accuracy and treatment efficiency. In contrast, AI technology, with its advanced data processing and analysis capabilities, offers a powerful alternative. AI can rapidly analyze vast amounts of client data to identify patterns and correlations, thereby assisting therapists in assessing patient symptoms more accurately and swiftly, and in identifying adverse emotions and cognitive distortions.

\subsubsection{Mental health assessment}

\paragraph{Disease diagnosis and symptom severity assessment}
AI is increasingly employed for the diagnosis and assessment of common mental illnesses such as depression and anxiety. 
~\cite{wanderley2022detection} proposed a methodology to support the diagnosis of mental disorders in psychiatric emergency context using vocal acoustic analysis and machine learning, targeting disorders like major depressive disorder, schizophrenia, bipolar disorder, and generalized anxiety disorder. 
~\cite{siddiqua2023aida} employed a survey questionnaire distributed among Bangladeshi students with a total of 684 responses were obtained, and utilized ten machine learning models and two deep learning models to predict three levels of depression severity (normal, moderate, and extreme) with remarkable accuracy.
~\cite{vuyyuru2023transformer}introduced the Trans-CNN, which effectively integrates the strengths of Transformer and Convolutional Neural Network (CNN) architectures. This model demonstrates superior performance in identifying depressive disorders compared to traditional CBT methods and standalone Transformer or CNN models by analyzing textual data such as patient narratives, treatment records, and diagnostic reports.

Despite these advancements, the existing research is centered around AI models focused on the diagnosis and assessment of singular psychological disorders, with limited research dedicated to the development of comprehensive models capable of simultaneously diagnosing multiple co-occurring psychological conditions.

\paragraph{Identifying cognitive distortions}

CBT postulates an individual's emotions and behaviors are influenced by their perception and interpretation of events. Cognitive distortions refer to erroneous or irrational patterns of thinking that can lead to negative emotions and maladaptive behaviors. During the initial assessment phase of CBT, therapists collaborate with clients to explore these cognitive patterns and reactions to specific events, aiming to assist the client in identifying and understanding potential cognitive distortions, which were categorized into ten types by Burns theory~\cite{burns1999feeling}. They include: all-or-nothing thinking, over generalization, mental filter, disqualifying the positive, jumping to conclusions, magnification and minimization, emotional reasoning, should statements, labeling and mislabeling, and blaming oneself or others.

Currently, AI technology facilitates the identification of cognitive distortions primarily through text classification techniques. Researchers utilize various textual data as inputs to build models and algorithms that automatically identify and categorize these cognitive distortions. However, due to the lack of publicly available structured datasets specifically designed for detecting cognitive distortions, researchers often turn to alternative data sources such as social media data~\cite{alhaj2022improving, wang2023cognitive}, personal blogs~\cite{simms2017detecting}, and everyday narratives~\cite{xing2017identifying, shickel2020automatic, mostafa2021automatic}. 
~\cite{shreevastava2021detecting} compared five classification algorithms for detecting cognitive distortions using a therapist Q\&A dataset obtained from Kaggle. ~\cite{tauscher2023automated} applied NLP methods to identify cognitive distortions from text messages exchanged between patients with severe mental illnesses and their clinical therapists.
Challenges in this task include dealing with short texts that lack contextual information and imbalanced data where certain distortion types are underrepresented, leading to poorer classification performance.
To address these issues, researchers have proposed various solutions. ~\cite{alhaj2022improving} suggest enriching short textual representations to improve cognitive distortions’ classification of the Arabic context over Twitter. It also utilizes a transformer-based topic modeling algorithm (BERTopic) that employs a pre-trained language model (AraBERT). ~\cite{ding2022improving} tackled data imbalance with approaches like data augmentation and domain-specific models, demonstrating the effectiveness of the pretrained language model MentalBERT.
Recent advances have expanded the scope of cognitive distortion detection to include multimodal datasets. This cross-modal research approach provides a more holistic perspective, enabling a more comprehensive detection of cognitive distortions. For instance, ~\cite{singh2023decode} proposed a multitask framework that integrates text, audio, and visual data to detect cognitive distortions, achieving significant performance improvements over existing state-of-the-art models.

Recently, LLMs have also shown promise in complex psychological tasks such as identifying cognitive distortions~\cite{qi2023evaluating, nazarova2023application, chen2023empowering, lim2024erd}.
~\cite{qi2023evaluating} conduct experiments to compare LLMs and supervised learning in cognitive distortion identification and suicide risk classification. The experimental results indicate that LLMs struggle with accurately identifying complex cognitive distortions in Chinese social media data, suggesting that deep learning algorithms remain the preferred solution for complex psychological tasks like cognitive distortion identification. 
~\cite{nazarova2023application} developed TeaBot, an AI tool fine-tuned on GPT-3, and employs CBT techniques to aid users in identifying and challenging distorted thoughts.
~\cite{chen2023empowering} introduced the Diagnosis of Thought (DoT) framework, which strategically prompts the LLMs to produce diagnosis rationales pertinent for cognitive distortions detection. Although the DoT method demonstrates its capability in classifying cognitive distortions, it also exhibits a notable flaw, namely, the model tends to over-diagnose cognitive distortions even when the user's statements are benign. 
\cite{lim2024erd} addressed this by introducing the ERD framework, involving extraction, reasoning, and debate among multiple LLM agents to classify cognitive distortions from user utterances. This approach significantly mitigates the issue of overdiagnosing cognitive distortions in the DoT method.

\paragraph{Emotion analysis}

Emotion analysis is essential for understanding individuals' emotional states. This understanding enables therapists or conversational agents to offer more empathetic support and feedback in CBT interventions, thereby strengthening the therapeutic alliance and enhancing the interactive experience~\cite{brave2005computers, provoost2019validating}. Furthermore, emotional states also serve as indicators of intrinsic goals, observable behaviors, and treatment efficacy. During the therapeutic process, individuals are more likely to benefit from treatment when they can effectively manage their emotions~\cite{mehta2021acceptability}. 
Therefore, during the initial assessment phase, therapists explore the clients' emotional responses and assist them acquire healthier emotion regulation strategies.
AI-assisted emotion analysis has become increasingly prevalent in mental health~\cite{tanana2021you, assunccao2022overview, khare2023emotion}.
~\cite{provoost2019validating} employed an emotion mining algorithm to assess the overall sentiment and five specific emotions expressed in texts written by patients during Internet-based cognitive behavioral therapy (ICBT), finding moderate agreement between the algorithm and human judgment in evaluating the overall sentiment, while the agreement was low for specific emotions.
~\cite{patel2019combating} proposed an intelligent social therapeutic chatbot. This chatbot defined several basic emotion labels, and based on these emotion labels, three deep learning algorithms were employed to extract emotions from user chat data. 
~\cite{kozlowski2023enhanced} introduced Terabot, a conversational system that enhanced sentiment and emotion recognition by integrating CBT techniques and replacing BERT with RoBERTa in a neural language model framework.
However, ~\cite{striegl2023deep} pointed out that some emotion recognition methods~\cite{fitzpatrick2017delivering, provoost2019validating} categorize emotions into discrete classes, failing to capture the continuum and diversity of emotions accurately. To address this, they proposed a deep learning-based dimensional text emotion recognition system within the context of CBT using the ALBERT pre-trained model~\cite{lan2019albert}, fine-tuned on emotion-annotated data for dimensional score prediction.

In recent years, ChatGPT, as a prominent example of conversational AI, shows potential in assisting individuals with emotion understanding and management.
~\cite{rathje2023gpt} initially assessed GPT's capability for overall sentiment analysis (positivity, negativity, or neutrality) in English and Arabic texts. They found that GPT demonstrates effective multilingual sentiment analysis, achieving performance levels comparable to top-performing machine learning models from previous years. Furthermore, they examined GPT's ability to accurately discern more nuanced discrete emotions such as anger, joy, fear, and sadness. The results demonstrated a high level of consistency between GPT's performance and human judgement.
~\cite{elyoseph2023chatgpt} also underscored GPT's superiority over humans in emotional cognition, as highlighted in their study.

Despite AI's capabilities in emotion detection and understanding, there remains a perception that individuals may perceive AI-driven support as lacking genuine emotional resonance compared to human interaction~\cite{yin2024ai}. Thus, future research should focus on making AI responses more conversational and human-like to address this perception. Additionally, to ensure rigor and utility in real clinical settings, human cross-validation is required.

\subsubsection{Personalized treatment selection: identifying CBT beneficiaries}

Within the realm of precision medicine, aligning clients with the most suitable treatment is a critical and significant objective~\cite{ozomaro2013personalized}. This involves selecting an effective treatment plan before commencing therapy, thereby avoiding the risks associated with ineffective treatments. CBT is a well-established psychotherapy approach to address mental health issues, however, its effectiveness varies among individuals experiencing different mental health conditions. 
This variability highlights the need to identify which clients are likely to respond favorably to CBT and to assess its effectiveness in treating specific mental health conditions. Researchers are increasingly exploring the potential of AI models to predict which clients may benefit from CBT and to determine the most appropriate treatment for each person~\cite{ball2014single, reggente2018multivariate, vieira2022can}. 
Studies suggest that AI can analyze extensive clinical data to forecast which patients are more likely to respond positively to CBT. This capability can assist clinicians in selecting suitable candidates for CBT, thereby optimizing resource allocation. 
For example, ~\cite{reggente2018multivariate} used machine learning with cross-validation to investigate the potential of functional connectivity patterns to predict obsessive-compulsive disorder (OCD) symptom severity in individuals after treatment. The findings have important clinical implications, particularly in the development of personalized medicine approaches aimed at identifying OCD patients who would derive the greatest benefit from intensive CBT. 
Similarly, ~\cite{vieira2022can} conducted a comprehensive review and quantitative synthesis of previous studies utilizing machine learning to predict the outcomes of CBT treatment in various diagnostic categories. They concluded that, on an individual level, the accuracy of predicting CBT benefit was approximately 74.0\%. 

Besides CBT, other types of psychological interventions, such as Psychodynamic Therapy (PDT) and person-centered counseling for depression may be equally effective. Moreover, not all clients may require high-intensity CBT interventions. Some individuals might benefit from low-intensity interventions. Traditionally, the trial-and-error method in psychotherapy can be both time-consuming and costly, with the risk of clients undergoing multiple ineffective treatments before finding the one that works best for them. This not only delays effective care but also increases the overall cost of treatment. Therefore, determining the most suitable treatment for specific individuals before the delivery of therapy is crucial. Addressing this gap, ~\cite{delgadillo2020targeted} and ~\cite{schwartz2021personalized} had applied machine learning algorithms to recommend optimal treatment approaches based on patients' pre-treatment characteristics. This method can contribute to substantial cost savings in mental health care, making high-quality psychotherapy more accessible and affordable.

Additionally, there is a growing interest in whether combining CBT with medication or other treatments can enhance outcomes. Several researchers have explored this possibility~\cite{gunlicks2020developing, pei2022acupuncture, stephenson2023comparing}. For example, ~\cite{gunlicks2020developing} found that combining CBT with medication is equally effective and more long-lasting than medication alone. Furthermore, they noted that the combination of medication and CBT could increase response rates and prolong the duration of effectiveness, especially when CBT is administered to clients responsive to pharmacotherapy.

These research underscores the potential of AI in refining the selection of psychological treatments and highlights the importance of personalized approaches in enhancing therapeutic outcomes.

\subsubsection{Client psychoeducation}
In CBT, client education is a crucial step both before and throughout the therapy process, as shown in Figure~\ref{fig:ai_cbt_Stages}. Before commencing therapy, psychoeducation involves introducing the fundamental principles and techniques of CBT, aiding them in understanding common patterns of negative emotions, thoughts, and behaviors and coping strategies ~\cite{mcginn2001allows}. In the treatment process, psychoeducation is reflected in teaching clients how to identify and assess automatic thoughts, how to perform cognitive restructuring, and how to find alternative explanations for automatic thoughts, thereby adjusting these automatic thoughts. By conveying that individuals can acquire and master skills to address psychological issues, psychoeducation enhances clients’ sense of self-efficacy and can even contribute to symptom reduction on its own~\cite{cuijpers1998psychoeducational}.

Psychoeducation can be delivered through various formats, including written materials, videos, audio recordings, as well as organizing specialized lectures and seminars. The advent of mobile applications has introduced novel avenues for delivering psychological education and support, capitalizing on their convenience and adaptability. Psychological education has been seamlessly integrated into several mobile applications, some of which have demonstrated efficacy in treating various mental disorders, including stress~\cite{rose2013randomized}, anxiety and depression~\cite{harrison2011mobile}, eating disorders\cite{cardi2013use}, post-traumatic stress disorder~\cite{reger2013pe}, and obsessive-compulsive disorder~\cite{whiteside2014case}. For instance, ~\cite{whiteside2014case} developed the smartphone application ``Mayo Clinic Anxiety Coach'', designed to deliver CBT for anxiety disorders, including OCD. This application comprises a psychological education module that instructs users on the utilization of the application, conceptualization of anxiety, descriptions of various anxiety disorders, explanations of CBT, and guidance on assessing alternative forms of therapy.   
Similarly, ~\cite{martinez2019assessment} developed a mobile-based embodied conversational agent (ECA) for the prevention and detection of suicidal behavior. The ECA introduces users to a segment of psychoeducational video in which a combination of cartoon-like images in motion are accompanied by a narrator explaining the basic elements of CBT. Users can access the video at any time through the application's main menu.
The advantages of delivering psychological education through mobile applications are self-evident: information becomes portable, and patients can access it easily.
\cite{heng2021rewind} integrated CBT within immersive gaming experiences to assist individuals suffering from generalized anxiety disorder (GAD). In this approach, CBT elements are seamlessly interwoven with the diegetic components of the game, providing psychoeducation in an engaging and entertaining manner.

AI-powered models facilitate personalized psychoeducation by tailoring modules to each patient's understanding and preferences. For example, Bhaumik et al.~\cite{bhaumik2023mindwatch} introduced MindWatch, a system harnessing AI-driven language models for early symptom detection and personalized psychological education. Within the psychological education module, they utilized the foundational Llama 2 model within the Amazon SageMaker Studio environment to deliver tailored education to individuals experiencing mental health issues.
Numerous chatbots incorporate modules for psychoeducation as well. For instance, ~\cite{jang2021mobile} developed Todaki, a chatbot for managing Attention Deficit Hyperactivity Disorder (ADHD). This chatbot offers tailored psychoeducation and brief CBT sessions, enabling individuals with ADHD to acquire self-help skills for managing their condition effectively. 
Similarly, ~\cite{su2022efficacy} created XIAO AN, an AI-assisted psychotherapy robot utilizing multi-modal signal recognition and natural interaction technology. XIAO AN monitors emotions and offers brief, comprehensive psychological therapy based primarily on CBT, integrating psychological education modules.

Through these approaches, integrating CBT principles into various technological platforms has facilitated the dissemination of psychological education, empowering individuals to better understand and manage their mental health.

\subsection{Integration of AI in CBT therapeutic process}

CBT intervention encompasses the implementation of various strategies, including cognitive restructuring, behavioral activation, and exposure therapy, etc. These strategies are intrinsically connected to cognition, emotions, and behavior. During CBT treatment, psychotherapists select appropriate strategies based on the patients' symptoms and needs to provide targeted treatment. In section~\ref{sec:strategies}, we will delve into how AI technology can enhance each CBT strategy. Following this, Section~\ref{sec:all_strategies} will focus on how researchers are developing chatbots to integrate multiple CBT strategies, offering comprehensive mental health support to users. Next, Section~\ref{sec:real_time_monitor} introduces how wearable devices or smartphone applications equipped with AI technology can now be used to monitor patients' psychological states during CBT treatment. Finally, Section~\ref{sec:outcome_predict} introduces some researchers utilize AI to make early predictions about the outcomes of CBT treatments.

\subsubsection{AI Augmentation of Therapeutic Strategies for CBT delivery} \label{sec:strategies}

\paragraph{Cognitive restructuring (CR)}

Cognitive restructuring (CR) is a fundamental component of CBT, characterized as a structured, goal-oriented, and collaborative intervention. It involves the patient and therapist working together to identify and correct irrational thoughts, evaluations, and beliefs by generating more adaptive alternative cognitive patterns~\cite{clark2013cognitive}. Thereby alleviating negative emotions, improving mental health, and promoting healthier behaviors and coping strategies. To some extent, it is considered analogous to Cognitive Reappraisal~\cite{shurick2012durable, zhan2024large}.
Recent advancements have leveraged AI models to conduct CR effectively. The majority of these efforts aim to provide evidence that AI language models can effectively engender reframed thinking in response to negative emotional situations, as well as to compare the efficacy of different models in CR.
For instance, ~\cite{de2021formulating} collected a small-scale CBT dataset via the crowd-sourcing platform Prolific. They validated Google’s T5 transformer and BERT model for their ability to transform initial negative cognition into more positive or realistic alternatives. Human evaluation revealed that T5-generated reconstructions resembled original human-written responses more closely, while BERT showed a relatively lower performance but higher positive sentiment.
~\cite{maddela2023training} introduced PATTERNREFRAME, a novel dataset that incorporates personas and classical unhelpful thought patterns, extending the reframing task to include the identification and generation of thoughts corresponding to a given persona and unhelpful pattern. They evaluated various language models using prompt-based and fine-tuning methods for their efficacy in identifying and reframing these thoughts.
Unlike ~\cite{de2021formulating} and ~\cite{maddela2023training}, who conceptualize CR tasks as a sentence rewriting task, ~\cite{xiao2024healme} emphasize client empowerment over reliance on therapist-driven solutions. They propose, Helping and Empowering through Adaptive Language in Mental Enhancement (HealMe), a model utilizing Large Language Models (LLMs) for CR through three steps: distinguishing between situations and thoughts for a rational perspective, generating alternative perspectives through brainstorming to alleviate negative thinking, and providing suggestions that recognize the client's efforts and promote positive action.

Previous research primarily focused on the perspective of holistic CR, attempting to address how language models can be utilized to generate CR. 
However, ~\cite{sharma2023cognitive, sharma2023facilitating} have begun to delve deeper into exploring various dimensions of CR, and examining people's preferences for particular types of restructuring. 
To be more specific, ~\cite{sharma2023cognitive} investigated how negative thinking can be restructured, how language models can be utilized to facilitate such restructuring, and which types of restructuring individuals experiencing negative thoughts prefer. They introduced a framework comprising seven linguistic attributes for reconstructing thoughts and trained a retrieval-enhanced in-context learning model to generate reconstructed thoughts. Additionally, they conducted a randomized field study on the Mental Health America (MHA) website.
In another work, ~\cite{sharma2023facilitating} designed and evaluated a CR tool based on human language model interaction. This system leverages language models to assist individuals in various steps of CR, including identifying cognitive traps within thoughts and selecting more actionable, empathetic, or personalized reconstruction suggestions when reconstructing negative thoughts. Furthermore, they demonstrated that language models can not only be utilized for generating reassessments but also aid individuals in enhancing their own reassessment abilities.
~\cite{wang2024cognitive} argue that while ~\cite{sharma2023cognitive, sharma2023facilitating} explored the enhancement of reframed thoughts within a single attribute in one generation, their efforts were limited in addressing multiple features. Consequently, they developed ReframeGPT, leveraging GPT-3 as an inference engine to generate and iteratively refine reframed thoughts across various features, aiming to achieve high-quality reframing.
~\cite{li2024skill} focused on aligning GPT-4's reconstructive thinking with human performance, enhancing model performance in CR tasks by understanding the differences between human and AI-generated reconstructions.

Identifying a patient’s automatic thoughts and emotions is crucial for effective CR. In traditional face-to-face CBT sessions, therapists refine identified automatic thoughts through methods such as Socratic questioning. However, when delivering CR in ICBT, accurately capturing thoughts and emotions poses challenges. 
Hence, ~\cite{furukawa2023harnessing} trained the T5 model to predict emotions associated with each automatic thought. They then compared these predictions with judgments from relevant experts, demonstrating the accuracy of T5 predictions. The application of T5 in iCBT platforms holds promise for achieving more efficient CR. 
~\cite{shidara2022automatic} and ~\cite{jiang2024ai} also have both underscored the significance of automatically identifying patients' cognitive distortions and emotions in CR. In response to this, corresponding efforts have been made. 
Particularly, ~\cite{jiang2024ai}, drawing on the ABCD model in CBT, employed the pre-trained model ERNIE 3.0 to construct a hierarchical text classification model for extracting automatic thoughts and emotions from user statements. Furthermore, they also validated the efficacy of LLMs in this task.

\paragraph{Behavioral activation (BA)}
Behavioral activation (BA) is a essential therapeutic strategy in CBT, focusing on addressing observable behaviors for intervention~\cite{petersen2016massachusetts}. In BA, therapists collaborate with clients to formulate specific goals and plans, encourage the modification of negative behavior patterns, gradually increase the frequency and diversity of positive activities, and help clients regain a sense of pleasure and meaningful engagement in daily life. 
~\cite{lewinsohn1974behavioral} and ~\cite{jacobson1996component} demonstrated that behavioral activation achieves outcomes comparable to full CBT and is comparable to pharmacotherapy~\cite{dimidjian2006randomized}. Individuals with mental disease may have problems with reduced daily activities, loss of interest, avoidance behavior and social avoidance. 
To facilitate the application of BA, researchers have developed innovative interventions. 

Innovative interventions have been developed to facilitate BA application. ~\cite{dahne2019pilot} developed a Spanish-language behavioral activation self-help mobile application called !`Apt{\'\i}vate!. Within this app, the user identifies individualized values and associated activities across five life areas, including relationships, daily responsibilities, recreation, career and education, and health. Evaluation results also demonstrated the feasibility and preliminary effectiveness of this app.
~\cite{rohani2020mubs} developed a mobile health system called MUBS (Mobile-based Behavioral Activation System) to support patients with depression through behavioral activation therapy. The system implements a unique content-based probabilistic recommendation algorithm to motivate patients to engage in positive behaviors by providing activity directories and personalized rewards, helping them build awareness of behavioral activation.

While these studies primarily focus on evaluating the efficacy of BA or developing applications to assist, prompt and motivate patients to actively engage in various activities, without explicitly linking AI to BA. Against this background, ~\cite{madhu2022activity} and ~\cite{rathnayaka2022mental} explore how AI technology can be utilized to enhance the effectiveness of behavioral activation.
~\cite{madhu2022activity} proposed a novel approach for activity recognition utilizing AI. They employed multi-modal data (combining speech and text) and utilized a BERT model for emotion recognition, along with logistic regression for sentiment detection. Subsequently, an activity classification model was employed wherein emotion and sentiment were integrated with keywords to accurately identify the appropriate activity for behavioral activation. This method achieved accuracies exceeding 80\% in emotion recognition, sentiment detection, and activity recognition tasks.
~\cite{rathnayaka2022mental} designed and developed a chatbot named Bunji using AI technology to provide emotional support, personalized behavior activation, and remote health monitoring. Participatory evaluation in a pilot study environment also demonstrated the practicality and effectiveness of the chatbot.

\paragraph{Exposure therapy (ET)}

Exposure therapy, a common technique in CBT, involves gradually exposing patients to anxiety-provoking or phobias-provoking situations to help them adapt and reduce their fear responses~\cite{abramowitz2013practice}. 
It is used to treat various psychological disorders, including specific phobias, panic disorder, post-traumatic stress disorder (PTSD), obsessive-compulsive disorder (OCD), and social anxiety disorder.
In recent years, virtual reality (VR) technology has emerged as a potent adjunct to traditional exposure therapy~\cite{jameel2020virtual}, leading to the development of Virtual Reality Exposure Therapy (VRET), validated by research for its effectiveness within CBT~\cite{scozzari2011virtual, maples2017use, orskov2022cognitive}.

While VRET may not yield superior results in terms of therapeutic efficacy compared to traditional exposure therapy within the CBT framework, it presents unique advantages like enhanced patient comfort, safety, real-time customization of exposure content, and simulation of personalized real-life scenarios. The integration of VR with CBT has led to the development of interactive applications and serious games to augment therapeutic interventions. 
~\cite{heng2021rewind} developed ReWIND, an RPG-styled serious game based on the ABCDE model of CBT. The design intention of ReWIND is to immerse players diagnosed with symptoms of Generalized Anxiety Disorder (GAD) in various anxiety-inducing situations within a virtual environment and provide them with constructive psychological interventions. The game aims to deliver psychological education to players in an engaging and interactive manner.
Similarly, ~\cite{giordano2022alter} emphasize that CBT is the most successful protocol in gambling disorder (GD) treatment, and they developed a serious virtual game called Alter Game, based on VRET, aimed at preventing relapse in patients with gambling disorder.
Additionally, ~\cite{michelle2014cbt} proposed an Android application named CBT Assistant, where patients can choose, set, and customize their own exposure fear ladders based on a generic template for graded exposure therapy. This feature distinguishes CBT Assistant from other CBT applications.
The fusion of VRET with AI technology further enhances its efficacy in addressing mental health challenges.
~\cite{aahs2020artificial} proposed and discussed the idea of utilizing Explainable Artificial Intelligence (XAI) to enhance CBT treatment for speech anxiety in VR settings.
~\cite{rahman2022towards} explore a range of machine learning models' performance on the task of arousal prediction using publicly available datasets. They propose a pipeline to address the model selection issue with various parameter configurations in the context of VRET.

\paragraph{Homework}

A CBT session typically lasts 45-60 minutes, which is often insufficient for many patients. Homework assignments bridge the gap between sessions, allowing patients to apply learned skills and enabling therapists to assess skill acquisition and maintenance~\cite{beck1979cognitive, lebeau2013homework}.
In the professional literature, homework assignments are frequently delineated as precise, structured therapeutic tasks upon during sessions, intended for completion between sessions. This process entails collaborative delineation of therapeutic objectives for the homework, determining pertinent activities or data collection as components of the homework, strategizing the practical execution of the homework, and subsequently reviewing the homework during subsequent sessions~\cite{kazantzis2010unresolved}.
Homework assignments may encompass a range of activities within each session, such as engaging with relevant materials, documenting thoughts and emotions, practicing specific skills or behaviors, or engaging in communication with others~\cite{tang2017supporting}. 

Research has found that clients who consistently complete homework derive greater benefits from interventions compared to those who complete little or no homework~\cite{burns2000does}, yet traditional paper-based CBT homework pose various impediments that can significantly undermine users' motivation to complete tasks as instructed. 
Non-compliance with homework is cited as one of the most common reasons for the failure of CBT treatments~\cite{helbig2004problems}, persisting as a prevalent issue in clinical practice. The prevalence of digital devices and the internet has enabled the transformation of traditional homework into digital formats, thereby enhance CBT homework compliance. However, there is a lack of guidelines for designing mobile phone apps tailored for this purpose. Consequently, ~\cite{tang2017supporting} proposes six essential features of an optimal mobile app aimed at maximizing CBT homework compliance, aiming to provide theoretical guidance for the development of such applications. 

Innovative approaches, such as the integration of traditional diary writing with mobile technology and LLM, offer promising solutions to enhance homework compliance. For example, ~\cite{peretz2023machine} developed a machine learning model capable of recognizing the presence of homework assignments during therapy sessions based on natural language dialogue between therapists and clients in real-world settings, as well as determining the type of homework assigned. Such advancements hold significant promise in bolstering therapists' ability to assign and monitor homework tasks, ultimately fostering enhancements in therapeutic outcomes.
~\cite{nepal2024contextual} integrated traditional diary writing with mobile technology and LLM to create a diary application with contextual awareness, named MindScape. Specifically, the application utilizes real-time analysis of behavioral data collected from smartphones and employs LLM to provide personalized, contextually relevant writing prompts. These prompts are designed to guide users in reflection and contemplation, facilitating the recording of their thoughts within daily life contexts. This innovative approach not only fosters a habit of regular self-reflection but also addresses the challenge of homework compliance in CBT.
During CBT for tinnitus alleviates the patients are typically assigned various homework tasks, including diary writing and self-monitoring. These homework assignments primarily consist of handwritten text data. However, analyzing this data can be extremely time-consuming for therapists, leading to decreased treatment efficiency. 
To address this issue, ~\cite{jeong2024advancing} proposed utilizing LLMs like GPT-2 to analyze the homework data of patients undergoing CBT. Their goal is to predict the Tinnitus Handicap Inventory (THI) scores from the homework, which can, in turn, predict the outcomes of CBT treatment, thereby enabling the selection of more personalized and effective treatment plans. Additionally, they compared the performance of the latest language models, particularly Google's T5 and Flan-T5, in predicting THI scores. Finally, they looked ahead the application of this research to monitor and predict the effectiveness of CBT treatment in patients with depression.

In summary, AI augments various CBT strategies by leveraging natural language processing and machine learning techniques, improving effectiveness and engagement.

\subsubsection{Developing AI tools for comprehensive delivery of CBT} \label{sec:all_strategies}

While CBT strategies are often discussed individually, in clinical practice, they are typically combined to address the specific needs of clients.
Recent advancements in AI have fostered the creation of digital tools that utilize the principles and techniques of CBT. These tools integrate multiple CBT strategies, thus providing a comprehensive approach to treatment. AI-powered chatbots, also referred to as conversational agents or relational agents exemplify such innovations, with chatbots being particularly notable. According to a review by Abd et al.~\cite{abd2019overview}, out of 17 chatbots offering psychotherapy, 10 were based on CBT. They are capable of mimicking the communication styles of human therapists by engaging users in interactive dialogues that include both verbal and non-verbal exchanges. This allows them to offer efficient, personalized, and readily accessible CBT treatment, available 24/7.
Pioneering chatbots like Woebot, Wysa, Tess, and Youper are emblematic of this trend. Their innovative approaches and effectiveness have garnered significant attention and recognition as effective psychological support systems based on CBT.
In recent years, in response to diverse needs, researchers have developed many new AI tools based on CBT principles~\cite{oliveira2021initial, rizea2022deep, rani2023mental, sabour2023chatbot, su2022efficacy, nwoye2024schizobot}.
As summarized in Table~\ref{tab:ai_cbt_tools}, we have outlined specific AI tools currently utilized in CBT interventions

\begin{longtable}{m{6pc}p{23pc}}
\caption{AI tools used in current CBT intervention} \label{tab:ai_cbt_tools} \\
\toprule
\textbf{AI tools} & \textbf{Description} \\
\midrule
\endfirsthead
\caption[]{AI tools used in current CBT intervention (continued)} \\
\toprule
\textbf{AI tools} & \textbf{Description} \\
\midrule
\endhead
\midrule
\endfoot
\bottomrule
\endlastfoot
\textbf{Woebot}\footnotemark[1] & Woebot is a chatbot that offers CBT-based therapy for depression and anxiety. It engages users in daily conversations, tracks their emotions, and introduces CBT concepts through short videos or interactive word games. Using decision trees and natural language processing, Woebot responds empathetically and provides helpful suggestions. It can also detect concerning language and directs users to external resources if needed. \\
\midrule

\textbf{Wysa}\footnotemark[2] & Wysa is a therapy chatbot designed to support mental health issues like depression, anxiety, stress, and loneliness. It utilizes CBT, mindfulness, and positive psychology techniques. Instead of AI-generated responses, Wysa uses pre-crafted therapeutic conversations developed by clinicians for safe and effective interactions. Its adaptive AI understands complex user inputs and offers empathetic feedback and tailored CBT-based tools.\\
\midrule

\textbf{Youper}\footnotemark[3] & Youper is a chatbot that delivers CBT through three steps: a personalized mental health assessment, instant support via conversations, and symptom monitoring. It uses a decision tree to select responses, conducts real-time emotion analysis, and provides CBT interventions based on the user's emotional state.\\
\midrule

\textbf{Tess}\footnotemark[4] & Tess is a psychological AI chatbot providing brief conversations for mental health support, psychoeducation, and reminders. It uses clinician-prepared statements to deliver interventions based on user-reported moods. Tess adjusts its responses based on user feedback, favoring CBT-based interventions for positive reactions and offering alternatives for neutral or negative ones. The platform is customizable to align with specific treatments or user demographics. \\
\midrule

\textbf{BetterHelp}\footnotemark[5] & BetterHelp is an online therapy platform that uses AI to match patients with licensed therapists and offers various approaches like CBT and psychodynamic therapy. \\
\midrule

\textbf{Rumi}\footnotemark[6]~\cite{oliveira2021initial} & Rumi is a  chatbot uses Rumination-focused Cognitive Behavioral Therapy (RFCBT) to explore the relationship between thoughts, feelings, and actions, aiming to improve mental health and reduce depressive and anxious symptoms.\\
\midrule

\textbf{Cloud Bot}~\cite{rizea2022deep} & Cloud Bot is a chatbot that utilized NLP technology to function as a psychologist. It focuses on applying a cognitive restructuring CBT technique to address users' issues. \\
\midrule

\textbf{Saarthi}~\cite{rani2023mental} & Saarthi is a chatbot using NLP and AI for delivering CBT and remote health monitoring to people with mental health issues. It offers real-time, evidence-based treatment that is accessible, affordable, and convenient, aiming to reduce anxiety and depression symptoms and enable long-term mental health monitoring. \\

\midrule
\textbf{SchizoBot}~\cite{nwoye2024schizobot} & SchizoBot is a chatbot using artificial neural networks to deliver CBT for managing schizophrenia, aiding clinicians and ensuring consistent therapy administration for patients.\\
\midrule

\textbf{XIAO AN}~\cite{su2022efficacy} & XIAO AN is a \textbf{Chinese} AI psychotherapy robot designed to monitor emotions and provide effective therapy, primarily using CBT principles. It has shown effectiveness in treating anxiety disorders in clinical trials without replacing therapists.\\
\midrule

\textbf{Emohaa}~\cite{sabour2023chatbot} & Emohaa is a \textbf{Chinese} conversational agent, which consists of two platforms: one is template-based (CBT-Bot) for structured conversations and exercises based on Cognitive Behavioral Therapy principles, while the other (ES-Bot) allows for open-ended discussions on emotional issues and provides emotional support. \\
\end{longtable}

\footnotetext[1]{\textbf{Woebot}: \url{https://woebothealth.com/}}
\footnotetext[2]{\textbf{Wysa}: \url{https://www.wysa.com/}}
\footnotetext[3]{\textbf{Youper}: \url{https://www.youper.ai/}}
\footnotetext[4]{\textbf{Tess}: \url{https://www.cass.ai/x2ai-home}}
\footnotetext[5]{\textbf{BetterHelp}: \url{https://www.betterhelp.com/}}
\footnotetext[6]{\textbf{Rumi}: \url{https://www.facebook.com/rumibot.bot/}}

By offering efficient, personalized, and readily accessible CBT treatment, these AI agents provide round-the-clock support to individuals undergoing therapy, which to some extent helps alleviate the pressure of insufficient medical resources. Nevertheless, it's worth noting that given the sensitivity of mental health support, it is crucial to assess the chatbots before presenting them to users.

\subsubsection{Real-time monitoring and feedback}\label{sec:real_time_monitor}

Traditional CBT often lacks real-time monitoring capabilities, relying solely on self-reported data from clients. However, the advent of mobile health technologies offers new possibilities for monitoring client health outside CBT sessions. According to research by ~\cite{yang2018iot}, CBT is well-suited for mobile platform applications, which facilitate real-time monitoring of user states and provide valuable data to therapists. Consequently, numerous mobile applications have been developed to deliver CBT treatment and track users' psychological states in real-time. 
~\cite{addepally2017mobile} noted that loneliness can be a risk factor for depression. To address this issue, the MoodTrainer application was developed, which tracks users' locations and isolating behaviors in real-time and provides CBT interventions when it detects relevant behaviors. 
Also, ~\cite{michelle2014cbt} introduced an Android application named CBT Assistant. This APP analyzes input data from individuals with social anxiety disorder (SAD) to identify stressors or situations triggering their mental health issues and assesses their severity. Additionally, some mobile applications are designed for specific CBT treatment scenarios, such as continuously tracking the sleep patterns of individuals with sleep disorders~\cite{schabus2023efficacy} or providing cessation monitoring and CBT examples to smokers~\cite{alsharif2015cognitive}, thereby enhancing their success rates. These mobile applications leverage the capabilities of mobile devices to collect real-time data and provide personalized interventions and support to individuals undergoing CBT treatment.

In recent years, wearable devices and smartphone applications equipped with AI technology are emerging as a trend for monitoring patients' psychological states. During CBT process, AI algorithms can monitor stress levels, physical activity, speech changes, and other indicators to detect variations in a patient's psychological state. These algorithms can send timely alerts to both patients and healthcare providers. 
~\cite{garcia2015automatic} embedded accelerometer sensors into smartphones to detect stress levels using classification algorithms like naive bayes and decision trees, achieving 71\% accuracy. Additionally, the long-term collection and analysis of large datasets can help therapists and patients gain a deeper understanding of the patient's mental health patterns. With the assistance of AI, therapists can access more comprehensive information, enabling them to discern trends in mental health, identify triggers, and evaluate treatment efficacy. These insights are crucial for making informed therapeutic decisions and designing personalized interventions. 
~\cite{goodwin2019predicting} collected physiological and movement data from wrist-worn biosensors in 20 adolescents diagnosed with ASD. They developed prediction models utilizing ridge-regularized logistic regression. These models demonstrate high accuracy in forecasting instances of aggressive behavior towards others occurring within the subsequent minute. Such advancements lay the groundwork for proactive behavioral interventions and timely adaptive intervention systems in the future.

Despite the growing use of AI-equipped wearable devices and mobile applications for monitoring psychological states and data collection, there is relatively limited research specifically focused on their application to CBT, particularly considering the unique demands of real-time interaction inherent in this therapeutic approach.

\subsubsection{Early prediction of treatment outcomes}\label{sec:outcome_predict}
Even among those for whom CBT proves effective, there may be a subset with limited benefits, potentially leading to a misallocation of both client time and limited healthcare resources. Hence, predicting treatment outcomes early in the therapeutic journey assumes critical importance, particularly to prevent the misallocation of client time and healthcare resources. Recent advancements in AI and large-scale data analysis have enabled the development of models capable of forecasting individual responses to CBT with high accuracy. 
~\cite{hahn2015predicting} applied a novel machine learning approach to predict individual-level response to CBT using fMRI data in patients diagnosed with panic disorder and agoraphobia (PD/AG). Similarly, ~\cite{tolmeijer2018using} utilze the machine learning methods to predict how people will respond when offered CBT for psychosis. Their two-step methodology involved first identifying potentially predictive regions and then developing a model based on these regions to make individual-level predictions. In a large-scale study, ~\cite{kaldo2021ai} analyzed data from over 6000 patients undergoing Internet-delivered CBT (ICBT). They evaluated the accuracy of various machine learning algorithms in predicting treatment outcomes and explored the integration of these algorithms into an Adaptive Treatment Strategy.
~\cite{isacsson2023machine} further examined the clinical utility of machine learning in predicting ICBT outcomes. They investigated the optimal timing within the treatment process for the model's predictive accuracy to support adaptive treatment strategies, proposing an optimal predictive model and offering specific recommendations based on their comprehensive analysis.
Despite advancements, traditional predictive models often exhibit limited accuracy, particularly in assessing the effectiveness of treating adolescent social anxiety. Under this background, ~\cite{zheng2022prediction} addressed this by employing deep learning techniques to construct a predictive model for the correlation between CBT and adolescent social anxiety, showcasing significantly improved predictive accuracy and reduced complexity compared to traditional models. 
~\cite{prasad2023deep} seeks to develop a state-of-the-art deep-learning framework for predicting clinical outcomes in ICBT by leveraging large-scale, high-dimensional time-series data of client-reported mental health symptoms and platform interaction data. 

Numerous researchers have also explored the use of specific treatment outcome predictors and potential biomarkers associated with particular diseases. For instance, ~\cite{wei2023towards} utilized the Hamilton Depression Rating Scale (HDRS) score as the primary outcome measure to investigate symptom changes in subjects undergoing CBT. Employing machine learning algorithms, they developed a support vector regression model, ultimately identifying left dorsolateral prefrontal cortex (DLPFC) Regional Homogeneity (ReHo) as a neuroimaging biomarker for the therapeutic effects of CBT in depression. Many previous studies have relied on highly selective samples to predict the outcomes of CBT. However, few have utilized routine available socio-demographic and clinical data to accomplish this task. 
Therefore, ~\cite{hilbert2020predicting, hilbert2021identifying} applied machine learning methods to clinical and socio-demographic data to predict the mental health treatment outcomes of individual patients. Their findings suggest that using routine data alone can feasibly predict treatment outcomes for mental disorders, with accuracy significantly surpassing chance levels.

These studies collectively highlight the promise of integrating advanced AI methodologies with clinical practice to enhance the early prediction of CBT treatment outcomes, offering potential pathways for more tailored and effective therapeutic interventions.

\subsubsection{Evaluation of therapists and clients in CBT treatment}
The application of AI in CBT extends beyond the diagnosis and assessment of disorders, it also encompasses crucial evaluations of therapists and clients by analyzing conversation data during therapy sessions.

\paragraph{Therapist treatment quality assessment}
Given the prevalence of mental health issues, ensuring the quality of psychotherapy is crucial to addressing the growing mental health demands and the complexities of the social environment. Traditionally, quality assessment is performed by human evaluators who listen to therapy recordings and review therapy notes to assess specific therapeutic skills. 
However, this approach is costly and time-consuming, resulting in limited feasibility and hindering widespread implementation in practical settings. To address these challenges, some researchers and technology developers have begun exploring the use of automated techniques to monitor the quality of psychotherapy. AI offers automated solutions for assessing therapy sessions and monitoring treatment fidelity of CBT sessions~\cite{chen2022leveraging}.
For example, ~\cite{ewbank2020quantifying} used a large-scale dataset containing session transcripts from more than 14000 patients receiving internet-enabled CBT (IECBT) to train a deep learning model to automatically categorize therapist utterances according to the role that they play in therapy, generating a quantifiable measure of treatment delivered. The closer the content provided by the therapist aligns with standard CBT protocols, the more positively it correlates with significant symptom improvement in patients. Conversely, the quantity of content unrelated to therapy shows a negative association. This method allows for the indirect assessment of the efficacy of CBT psychotherapy provided by the therapist.
However, ~\cite{flemotomos2021automated} emphasize that, for CBT, the most commonly utilized coding scheme is the Cognitive Therapy Rating Scale (CTRS), which defines a set of 11 session-level codes reflecting skills and techniques specific to the intervention. Consequently, they introduced a model for quality assessment of psychotherapy sessions based on adapted BERT representations of therapy language use. Their analysis focused on the binary classification of CBT sessions concerning the overall CTRS score. ~\cite{chen2022automated} also utilized CTRS scores to assess the quality of CBT sessions. However, unlike ~\cite{flemotomos2021automated}, they proposed a hierarchical framework for automatically evaluating the quality of transcribed CBT interactions. ~\cite{ardulov2022local} proposed an approach that explicitly focuses on control-affine dynamical system models. They attempted to extract local dynamic modes from short windows of conversation and learn to correlate the observed dynamics with CBT competence.
Furthermore, some studies aim to identify areas where the therapist excels and areas where improvement is needed by analyzing recordings of therapy sessions, comparing therapists' language and behavior against standards of specific therapeutic models~\cite{stirman2021novel, flemotomos2022automated, zhang2023you, wang2024patient}. This process aids therapists in enhancing their professional skills, thereby improving the overall quality of therapy.
Particularly, ~\cite{wang2024patient} developed PATIENT-${\Psi}$, a novel patient simulation framework for CBT training. Specifically, they constructed diverse patient profiles and corresponding cognitive models based on CBT principles, and used a large language model to act as a simulated therapy patient. This role-playing therapeutic scenario helps mental health trainees practice CBT skills.

\paragraph{Predictive analysis of clients treatment adherence}
Treatment adherence refers to the extent to which clients actively engage in and comply with the advice and instructions provided by healthcare professionals, thereby adhering to the treatment regimen, and it directly impacts the effectiveness and outcomes of the treatment~\cite{dimatteo2002patient}. Therefore, client adherence evaluation is essential components of effective healthcare delivery. Particularly, against the backdrop of ICBT, there has been a reduction in face-to-face interactions between healthcare professionals and clients, which may lead to low client engagement and high dropout rates. AI models can analyze client behavior and responses during therapy sessions to evaluate their understanding, engagement, and adherence to treatment content. 
~\cite{cote2022adherence} presents a minimally data-sensitive approach, based on a self-attention deep neural network, to perform adherence forecasting of clients undergoing G-ICBT. This study leverages between 7 to 42 days of user-interaction data (login/logout) from the eMeistring platform. 
This analysis can identify individuals who may be at risk of early dropout from intervention. With this information, clinicians can implement meaningful and targeted interventions such as reminders, scheduling direct interactions, or modifying the treatment approach to prevent premature termination.

\subsection{Integration of AI in Post-treatment Stage}
\subsubsection{Long-term treatment response and relapse prediction} 

Since current CBT interventions typically have relatively short durations, leading to inadequate durability of effects for some patients post-treatment, with a considerable proportion experiencing relapse after treatment cessation. Hence, predicting individual long-term clinical responses and relapse risks remains a significant challenge. AI presents a promising avenue for predicting relapse risks and enhancing the durability of CBT's effects in patients undergoing CBT. 
~\cite{maansson2015predicting} proposed the first study using the multivariate SVM–fMRI method to successfully predict long-term treatment response of ICBT for social anxiety disorder, achieving a predictive accuracy reached 92\% (95\% confidence interval 73.2–97.6).
~\cite{lorimer2021dynamic} utilized machine learning techniques, specifically the XGBoost algorithm, to analyze follow-up data of patients undergoing Low-Intensity Cognitive Behavioral Therapy (LiCBT). They developed a dynamic prediction tool capable of identifying cases with higher relapse risk at four distinct time points during the patients' treatment journey, with the ability to dynamically adjust prediction accuracy. This tool's capacity for early identification of high-relapse cases, combined with targeted relapse prevention measures, can significantly enhance the long-term efficacy of LiCBT in situations where psychological resources are limited.

In conclusion, the integration of AI in CBT holds promise for revolutionizing the delivery and efficacy of psychological interventions. By leveraging AI technologies throughout various stages of the treatment process, clinicians can enhance treatment outcomes, personalize interventions, and optimize resource allocation, ultimately improving mental health outcomes for individuals on a global scale.

\section{Datasets} \label{sec:datasets}

Datasets play a crucial role in driving research at the intersection of CBT and AI, providing foundational material for training, testing, and validating AI algorithms, and furnishing practitioners and researchers with the bedrock for constructing and evaluating CBT models. This section aims to review existing publicly datasets relevant to the application of CBT and AI. 
In the context of disease diagnosis and assessment tasks, numerous datasets have already been extensively reviewed in various comprehensive articles. Consequently, we will not elaborate on these datasets further in this paper. Moreover, it is worth noting that for the task of selecting personalized treatment strategies, there is a noticeable lack of publicly available datasets. This paper will not cover these datasets in detail. Instead, we focus on datasets for specific CBT-related tasks such as identifying and classifying cognitive distortions, conducting cognitive restructuring, and analyzing CBT conversation data. For clarity, datasets where descriptions are unclear or ambiguous in the literature will not be discussed in this paper.

Table~\ref{tab:simplified_version_of_cd_dataset} provides a comprehensive overview of datasets used for detecting and classifying cognitive distortions. As can be seen from this table, most of the cognitive distortion data sets are in English, and fewer are in Chinese. Moreover, these datasets present several notable challenges: First, low reliability. The labeling criteria vary significantly across different datasets, leading to inconsistencies. Moreover, the inherently subjective nature of labeling cognitive distortions further compromises reliability. Second, there is a pronounced issue of data imbalance, with certain classes of cognitive distortions lacking adequate representation. This imbalance hampers the model's ability to generalize well across all classes.
Table~\ref{tab:cognitive_restructuring_dataset} summarizes datasets relevant to cognitive restructuring. Additionally, Table~\ref{tab:cbt_session_dataset} outlines datasets that involve CBT conversations. For these areas, high-quality, annotated datasets are particularly scarce for both cognitive restructuring and CBT conversation analysis. 

\begin{longtable}{p{7pc}p{22pc}}
\caption{Cognitive Distortions Dataset. For a clean presentation, we have only shown the simplified version of the cognitive distortion dataset, see Appendix~\ref{sec:appendix_dataset} for more details.} 
\label{tab:simplified_version_of_cd_dataset} \\
\toprule
\textbf{Study} & \textbf{Description} \\
\midrule
\endfirsthead
\caption[]{\raggedright Cognitive Distortions Dataset (continued)} \\
\toprule
\textbf{Study} &  \textbf{Description} \\
\midrule
\endhead
\midrule
\endfoot
\bottomrule
\endlastfoot

~\cite{wang2023cognitive}  & 
\makecell[{{p{22pc}}}]{
\textbf{Size of Dataset}: A total of 1644 data entries across 11 types of cognitive distortions, and 2000 entries for normal cases. \\ 
\textbf{Language}: Not reported. \\
\textbf{Data Modalities}: Text. 
} \\
\midrule

~\cite{elsharawi2024c} & 
\makecell[{{p{22pc}}}]{
\textbf{Size of Dataset}: A total of 34370 samples across 14 types of cognitive distortions.\\ 
\textbf{Language}: English. \\
\textbf{Data Modalities}: Text. 
} \\
\midrule

~\cite{shickel2020automatic} & 
\makecell[{{p{22pc}}}]{
\textbf{Size of Dataset}: Dataset CrowdDist contained 7,666 texts across all 15 distortions, with an average of 511 responses per distortion. Dataset MH contained two subsets: MH-C was annotated with 15 cognitive distortion labels with 1164 distorted texts, and the MH-D dataset was annotated with binary distorted/non-distorted labels, distorted for 1605 texts, not distorted for 194 texts. \\ 
\textbf{Language}: English. \\
\textbf{Data Modalities}: Text. 
} \\
\midrule

~\cite{lim2024erd} & 
\makecell[{{p{22pc}}}]{
\textbf{Size of Dataset}: A total of 2530 samples across 10 types of cognitive distortions. \\ 
\textbf{Language}: English. \\
\textbf{Data Modalities}: Text. \\
} \\
\midrule

~\cite{shreevastava2021detecting} & 
\makecell[{{p{22pc}}}]{
\textbf{Size of Dataset}: A total of 3000 samples, 39.2\% were marked as not distorted, while the remaining were identified to 10 type of distortions. \\ 
\textbf{Language}: English. \\
\textbf{Data Modalities}: Text. 
} \\
\midrule

~\cite{de2021formulating} & 
\makecell[{{p{22pc}}}]{
\textbf{Size of Dataset}: A total of 200 samples across 14 types of cognitive distortions. \\ 
\textbf{Language}: English. \\
\textbf{Data Modalities}: Text. \\
} \\
\midrule

~\cite{sharma2023cognitive} & 
\makecell[{{p{22pc}}}]{
\textbf{Size of Dataset}: A total of 1077 samples across 13 types of cognitive distortions. \\ 
\textbf{Language}: English. \\
\textbf{Data Modalities}: Text. 
} \\
\midrule

~\cite{maddela2023training} & 
\makecell[{{p{22pc}}}]{
\textbf{Size of Dataset}: About 10k samples across 10 types of cognitive distortions. \\ 
\textbf{Language}: English. \\
\textbf{Data Modalities}: Text. 
} \\
\midrule

~\cite{wang2023c2d2} & 
\makecell[{{p{22pc}}}]{
\textbf{Size of Dataset}: 7,500 cognitive distortion thoughts across 7 types of common cognitive distortions. \\ 
\textbf{Language}: \textbf{Chinese}. \\
\textbf{Data Modalities}: Text. 
} \\
\midrule

~\cite{qi2023evaluating} & 
\makecell[{{p{22pc}}}]{
\textbf{Size of Dataset}: A total of 3407 posts across 12 types of cognitive distortions. \\ 
\textbf{Language}: \textbf{Chinese}. \\
\textbf{Data Modalities}: Text. 
} \\
\midrule

~\cite{na2024cbt} & 
\makecell[{{p{22pc}}}]{
\textbf{Size of Dataset}: A total of 22,327 samples across 10 types of cognitive distortions. \\ 
\textbf{Language}: \textbf{Chinese}. \\
\textbf{Data Modalities}: Text.
} \\

\botrule

\end{longtable}

\begin{table}[!hbtp]
\centering
\caption{Cognitive Restructuring Dataset.}
\label{tab:cognitive_restructuring_dataset}
\begin{tabular}{m{3pc}>{\centering\arraybackslash}m{6pc} >{\centering\arraybackslash}m{6pc} >{\centering\arraybackslash}m{12pc}}
\toprule
\textbf{Study} & \textbf{Tasks} & \textbf{Dataset source}  & \textbf{Description}  \\
\midrule

~\cite{sharma2023cognitive} & Cognitive Reframing of Negative Thoughts through Human-Language Model Interaction. & Thought Records Dataset and Mental Health America (MHA) website. & 
\makecell[{{p{12pc}}}]{
\textbf{Size of Dataset}: A total of 300 situations-thoughts pairs, reframed thoughts per situation. \\ 
\textbf{Component}: situation; thought; reframe; thinking\_traps\_addressed. \\
\textbf{Language}: Chinese. \\
\textbf{Data Modalities}: Text. \\
\textbf{Open Source}: Yes\footnotemark[1].
} \\
\noalign{\vskip 1em} 

~\cite{lin2024detection} & Detection of Cognitive Distortion and cognitive restructuring. & The corpus labeling utilizes one specialized open\-source dataset, the Chinese psychological Q\&A dataset PsyQA~\cite{sun2021psyqa}. & 
\makecell[{{p{12pc}}}]{
\textbf{Size of Dataset}: A total of 1900 sentences. \\ 
\textbf{Component}: original text and reconstruction text. \\
\textbf{Language}: Chinese. \\
\textbf{Data Modalities}: Text. \\
\textbf{Open Source}: Yes\footnotemark[2].
} \\
\noalign{\vskip 1em} 

~\cite{shidara2022automatic} & Identification and evaluation of automatic thoughts and cognitive restructuring. & Recruit participants. & 
\makecell[{{p{12pc}}}]{\textbf{Size of Dataset}: Not reported. \\ 
\textbf{Language}: Japanese. \\
\textbf{Data Modalities}: Text. \\
\textbf{Open Source}: Yes. It can be obtained by sending an email.
} \\

\botrule
\end{tabular}

\end{table}
\footnotetext[1]{~\cite{sharma2023cognitive}: \url{https://github.com/behavioral-data/Cognitive-Reframing}}
\footnotetext[2]{~\cite{lin2024detection}: \url{https://github.com/405200144/Dataset-of-Cognitive-Distortion-detection-and-Positive-Reconstruction}}

\begin{table}[!hbtp]
\centering
\caption{CBT session dataset.}
\label{tab:cbt_session_dataset}

\begin{tabular}{m{3pc}>{\centering\arraybackslash}m{6pc} >{\centering\arraybackslash}m{6pc} >{\centering\arraybackslash}p{12pc}}
\toprule
\textbf{Study} & \textbf{Tasks} & \textbf{Dataset source}  & \textbf{Description}  \\
\midrule
~\cite{lee2023chain} & Enhancing Empathetic Response of Large Language Models Based on Psychotherapy Models. & Crowdsourced Reddit posts of mental health from~\cite{sharma2020computational}. & 
\makecell[{{p{12pc}}}]{\textbf{Size of Dataset}: Three levels of empathy strategies, and the number of pairs for each strategy was as follows: ``emotion reaction''=1,047, ``exploration''=481, and ``interpretation''=1,436. \\ 
\textbf{Language}: English. \\
\textbf{Data Modalities}: Text. \\
\textbf{Open Source}: Yes\footnotemark[1]. \\ 
} \\
\noalign{\vskip 1em}

~\cite{na2024cbt} & Enhance the precision and efficacy of psychological support through LLMs & Get PsyQA Questions~\cite{sun2021psyqa}, which is derived from the Chinese online mental health support forum Yixinli, then utilizes CBT Prompt to generate CBT answers. & 
\makecell[{{p{12pc}}}]{\textbf{Size of Dataset}: 22,327 entries, each comprising questions, descriptions, and CBT responses. \\ 
\textbf{Language}: Chinese. \\
\textbf{Data Modalities}: Text. \\
\textbf{Open Source}: Yes. Follow the data copyright protocols and obtain it by email. \\ 
\textbf{Note}: The questions in the dataset originate from online mental health forum, and the responses are generated by ChatGPT, not professionals.
} \\

\botrule
\end{tabular}
\end{table}
\footnotetext[1]{~\cite{lee2023chain}: \url{https://github.com/behavioral-data/Empathy-Mental-Health}}

\section{Discussion} \label{sec:discussion}

The integration of AI into CBT has led to significant advances in pre-treatment assessment, the therapeutic process, and post-treatment follow-up. First, AI has improved efficiency by assisting with pre-treatment screening and diagnosis, reducing therapist workload, and enabling real-time adjustments to treatment plans through predictive capabilities. Second, AI has enabled more personalized CBT by analyzing rich patient data and identifying subtle patterns, enabling customized treatment plans that go beyond the traditional reliance on therapist expertise alone. Finally, AI-powered CBT platforms have increased the accessibility of mental health services by providing remote and cost-effective treatment options through online platforms and mobile applications that offer 24/7 support.
However, there are several limitations to current AI applications in CBT. In pre-treatment assessment, while AI excels in diagnosing psychological disorders and assessing cognitive distortions and emotional states using textual data, it has fewer applications in analyzing video, audio, and behavioral data. The potential of multimodal data for comprehensive diagnosis remains largely untapped. During the therapeutic process, AI enhances individual CBT strategies but struggles to cover the complexity of interventions comprehensively. AI tools like chatbots and virtual therapy assistants vary in quality and lack standardized development, leading to inconsistencies. Moreover, there are inadequate metrics for evaluating feasibility, engagement, and satisfaction, complicating platform comparison and improvement. Most AI-driven wearable devices and mobile applications focus on general indicators like heart rate and activity levels, lacking specialized tools for monitoring CBT-specific indicators. In post-treatment, while AI aids in predicting long-term treatment responses and relapse risks, challenges include variability in prediction data and insufficient research on utilizing predictive results to formulate personalized intervention strategies.

The future of AI in CBT holds great potential. Autonomous learning and adaptive therapy systems can emulate human CBT therapists, engaging in multi-round interactions with patients and adjusting strategies based on real-time feedback. Group intelligence support and decision-making systems can improve both therapist guidance and patient outcomes by aggregating the experience of experienced practitioners and facilitating intelligent social support. Cognitive augmentation and assistive systems can develop personalized tools to enhance cognitive function, thereby increasing the effectiveness of CBT. Customized, personalized CBT models can adapt to users' specific data, such as social background, culture, education, and environment, to provide tailored responses and interventions that increase therapeutic effectiveness and user satisfaction.
Despite the potential of AI, several challenges need to be addressed. Data security and privacy are paramount, requiring compliance with privacy regulations, anonymization of sensitive information, and advanced encryption techniques. Ethical and algorithmic bias must be mitigated by ensuring data diversity, involving multiple stakeholders in development, and continuously monitoring AI systems. Model explainability and transparency are essential for responsible AI decisions, requiring methods to improve interpretability and the use of rigorously tested models. Over-reliance on AI is a risk because the success of CBT depends on the therapist-patient relationship, which AI cannot replicate. AI should be a complementary tool, not a replacement. Evaluation of models in clinical practice is necessary to assess real-world effectiveness, acceptance, trust, and usability, taking into account training costs and impact on medical practice.
In summary, while AI offers promising enhancements to CBT, it must be used responsibly and ethically, complementing the guidance of professional therapists. The ultimate goal is to use technology to support, not replace, human-centered mental health care.

\section{Conclusion} \label{sec:conclusion}

In this paper, we have conducted a comprehensive literature review of the integration of AI technology into CBT . We explored the application of AI throughout the CBT process, highlighting its significant transformative impact and existing limitations. Subsequently, We have summarized publicly available datasets relevant to various CBT-related tasks to provide a foundation for future research. We suggested future research directions and acknowledged the practical challenges that AI faces in clinical settings.
Overall, our review illuminates the multifaceted integration of AI in CBT, highlighting its potential while providing a nuanced understanding of its capabilities. We hope that the findings will guide future research, bring new perspectives to clinical practice, and contribute to the advancement of mental health care.

\backmatter

\bmhead{Acknowledgements}
This work was supported by grants from the National Natural Science Foundation of China (grant numbers:72174152, 72304212 and 82071546), Wuhan University Innovation and Entrepreneurship Projects for College Students (No: 202410486100), Fundamental Research Funds for the Central Universities (grant numbers: 2042022kf1218; 2042022kf1037), and the Young Top-notch Talent Cultivation Program of Hubei Province.
Guanghui Fu is supported by a Chinese Government Scholarship provided by the China Scholarship Council (CSC).

\section*{Declarations}

\begin{itemize}
\item Competing Interests: The authors have no competing interests to declare that are relevant to the content of this article.

\end{itemize}

\noindent

\newpage
\begin{appendices}

\section{Cognitive distortion dataset}\label{sec:appendix_dataset}
In this section, we provide full detail of the dataset as described in Section~\ref{sec:datasets}.

\begin{longtable}{m{4pc}>{\centering\arraybackslash}m{5pc} >{\centering\arraybackslash}m{6pc} >{\centering\arraybackslash}p{15pc}}
\caption{Cognitive Distortions Dataset.}
\label{tab:cognitive_distortions_dataset} \\
\toprule
\textbf{Study} & \textbf{Tasks} & \textbf{Dataset source}  & \textbf{Description} \\
\midrule
\endfirsthead
\caption[]{Cognitive Distortions Dataset (continued)} \\
\toprule
\textbf{Study} & \textbf{Tasks} & \textbf{Dataset source} &  \textbf{Description} \\
\midrule
\endhead
\midrule
\endfoot
\bottomrule
\endlastfoot
~\cite{wang2023cognitive} & Cognitive distortion classification. & Published research papers on cognitive distortions, examples from articles on the web that introduce cognitive distortions and social media posts. & 
\makecell[{{p{15pc}}}]{
\textbf{Size of Dataset}: Before expansion: There were a total of 353 data entries for all cognitive distortions, and 1000 entries for normal cases. After expansion: There are now a total of 1644 data entries across 11 types of cognitive distortions, and 2000 entries for normal cases. \\ 
\textbf{Label}: 12 labels, including 11 cognitive distortions and no cognitive distortions. \\
\textbf{Language}: Not reported. \\
\textbf{Data Modalities}: Text. \\
\textbf{Open Source}: Yes.  \textbf{However, the article says it will share, but no link is given.  }
} \\
\noalign{\vskip 1em}

~\cite{shickel2020automatic} & Detection and Classification of Cognitive Distortions. & Dataset CrowdDist, comes from the popular crowdsourcing platform Mechanical Turk(MTurk). Dataset MH, comes from TAO Connect, an online mental health therapy service. & 
\makecell[{{p{15pc}}}]{
\textbf{Size of Dataset}: Dataset CrowdDist contained 7,666 text responses across all 15 distortions, with an average of 511 responses per distortion. Dataset MH contained two subsets: The MH-C dataset was annotated with 15 cognitive distortion labels with 1164 distorted texts, and the MHD dataset was annotated with binary distorted/non-distorted labels, distorted for 1605 texts, not distorted for 194 texts. \\ 
\textbf{Label}: situation; thought; reframe; thinking\_traps\_addressed.  \\
\textbf{Language}: English. \\
\textbf{Data Modalities}: Text. \\
} \\
\noalign{\vskip 1em} 

~\cite{elsharawi2024c} & Detection and Classification of Cognitive Distortions. & Facebook Empathetic Data, Twitter Data and Crowd-Sourcing. & 
\makecell[{{p{15pc}}}]{
\textbf{Size of Dataset}: A total of 34370 samples. \\ 
\textbf{Label}: 14 labels. \\
\textbf{Language}: English. \\
\textbf{Data Modalities}: Text. \\
\textbf{Open Source}: Yes. \textbf{However, the article says it will share, but no link is given.}
} \\
\noalign{\vskip 1em} 

~\cite{lim2024erd} & Detection and Classification of Cognitive Distortions. & Dataset Therapist Q\&A comes from crowd-sourced data science repository, Kaggle. & 
\makecell[{{p{15pc}}}]{
\textbf{Size of Dataset}: A total of 2530 samples. \\ 
\textbf{Label}: 10 labels. \\
\textbf{Language}: English. \\
\textbf{Data Modalities}: Text. \\
\textbf{Open Source}: Yes\footnotemark[1]. 
} \\
\noalign{\vskip 1em} 

~\cite{shreevastava2021detecting} & Automatic Detection and Classification of Cognitive Distortions. & Dataset Therapist Q\&A comes from crowd-sourced data science repository, Kaggle. & 
\makecell[{{p{15pc}}}]{
\textbf{Size of Dataset}: A total of 3000 samples, 39.2\% were marked as not distorted, while the remaining were identified to have some type of distortion. \\ 
\textbf{Label}: 11 labels,  including 10 cognitive distortions and no cognitive distortions. \\
\textbf{Language}: English. \\
\textbf{Data Modalities}: Text. \\
\textbf{Open Source}: Yes. \textbf{However, the article says it will share, but no link is given.}
} \\
\noalign{\vskip 1em} 

~\cite{de2021formulating} & Convert negative or distorted thoughts into more realistic alternatives. & A variety of sources such as CBT books, forums and public content aggregators. & 
\makecell[{{p{15pc}}}]{
\textbf{Size of Dataset}: A total of 200 samples. \\ 
\textbf{Label}: 14 labels. \\
\textbf{Language}: English. \\
\textbf{Data Modalities}: Text. \\
\textbf{Open Source}: Yes\footnotemark[2].  
} \\
\noalign{\vskip 1em} 

~\cite{sharma2023cognitive} & Cognitive Reframing of Negative Thoughts through Human-Language Model Interaction. & Thought Records Dataset and Mental Health America (MHA) website.  & 
\makecell[{{p{15pc}}}]{
\textbf{Size of Dataset}: A total of 1077 samples. \\ 
\textbf{Label}: 13 labels. \\
\textbf{Language}: English. \\
\textbf{Data Modalities}: Text. \\
\textbf{Open Source}: Yes\footnotemark[3].  
} \\
\noalign{\vskip 1em}

~\cite{maddela2023training} & Training models to generate, recognize, and reframe unhelpful thoughts. & Obtain a diversity of contexts, situations and thoughts from PERSONA\-CHAT dataset~\cite{zhang2018personalizing}, and ask crowdworkers to rewrite them. & 
\makecell[{{p{15pc}}}]{
\textbf{Size of Dataset}: About 10k examples of thoughts containing unhelpful thought. \\ 
\textbf{Label}: 10 labels. \\
\textbf{Language}: English. \\
\textbf{Data Modalities}: Text. \\
\textbf{Open Source}: Yes\footnotemark[4]. \textbf{However, this article says it will be shared and gives a github link, but it does not contain data.}
} \\
\noalign{\vskip 1em} 

~\cite{wang2023c2d2} & Cognitive Distortion Detection and investigate the association between cognitive distortions and mental health. & Carefully select and train volunteers to observe scenes and note possible cognitive distortions, then have experts evaluate the results. & 
\makecell[{{p{15pc}}}]{
\textbf{Size of Dataset}: 7,500 cognitive distortion thoughts. \\ 
\textbf{Label}: 7 common cognitive distortions. \\
\textbf{Language}: Chinese. \\
\textbf{Data Modalities}: Text. \\
\textbf{Open Source}: Yes\footnotemark[5]. 
} \\
\noalign{\vskip 1em}

~\cite{qi2023evaluating} & Classification of Cognitive Distortions. & Comments on a Sina Weibo post by the user ”Zoufan”. & 
\makecell[{{p{15pc}}}]{
\textbf{Size of Dataset}: A total of 3407 posts. \\ 
\textbf{Label}: 12 labels. \\
\textbf{Language}: Chinese. \\
\textbf{Data Modalities}: Text. \\
\textbf{Open Source}: Yes\footnotemark[6]. 
} \\
\noalign{\vskip 1em} 

~\cite{na2024cbt} & Enhance the precision and efficacy of psychological support through LLMs. & Get PsyQA Questions~\cite{sun2021psyqa}, which is derived from the Chinese online mental health support forum Yixinli, then utilizes CBT Prompt to generate CBTanswers. & 
\makecell[{{p{15pc}}}]{
\textbf{Size of Dataset}: A total of 22,327 samples. \\ 
\textbf{Label}: 10 labels. \\
\textbf{Language}: Chinese. \\
\textbf{Data Modalities}: Text. \\
\textbf{Open Source}: Yes. Follow the data copyright protocols and obtain it by sending an email. \\
\textbf{Note}: The questions in the dataset originate from online mental health forum, and the responses are generated by ChatGPT, not professionals. 
} \\
\noalign{\vskip 1em} 

\end{longtable}
\footnotetext[1]{~\cite{lim2024erd}: \url{https://www.kaggle.com/datasets/sagarikashreevastava/cognitive-distortion-detetction-dataset}}
\footnotetext[2]{~\cite{de2021formulating}: \url{https://github.com/itoledorodriguez/cbt-dataset}}
\footnotetext[3]{~\cite{sharma2023cognitive}: \url{https://github.com/behavioral-data/Cognitive-Reframing}}
\footnotetext[4]{~\cite{maddela2023training}: \url{https://github.com/facebookresearch/ParlAI/tree/main/projects/reframe_thoughts}}
\footnotetext[5]{~\cite{wang2023c2d2}: \url{https://github.com/bcwangavailable/C2D2-Cognitive-Distortion}}
\footnotetext[6]{~\cite{qi2023evaluating}:\url{https://github.com/HongzhiQ/SupervisedVsLLM-EfficacyEval}}

\newpage
\section{Abbreviation}\label{sec:appendix_B}

In this section, as shown in Table~\ref{tab:abbreviations}, we summarize the full names and abbreviations of various specialized terms mentioned throughout the text. By providing this list of abbreviations, our aim is to assist readers in quickly referencing and understanding the meanings of these terms, thereby enhancing comprehension of the content and its context within the paper.

\begin{table}[!htbp]
\caption{Full name and abbreviation of terms} \label{tab:abbreviations}
\centering
\begin{tabular}{ll} 
\toprule
Abbreviation & Full Name                                    \\ 
\midrule
ADHD         & Attention Deficit Hyperactivity Disorder     \\
AG           & Agoraphobia                                  \\
AI           & Artificial Intelligence                      \\
AR           & Augmented Reality                            \\
BA           & Behavioral Intervention                      \\
CBT          & Cognitive Behavioral Therapy                 \\
CCBT         & Computer-Based Cognitive Behavioral Therapy  \\
CD           & Cognitive Distortions                        \\
CNN          & Convolutional Neural Network                 \\
CR           & Cognitive Restructuring                      \\
CTRS         & Cognitive Therapy Rating Scale               \\
DLPFC        & Dorsolateral Prefrontal Cortex               \\
DoT          & Diagnosis of Thought                         \\
ECA          & Embodied Conversational Agent                \\
ET           & Exposure Therapy                             \\
GAD          & Generalized Anxiety Disorder                 \\
GD           & Gambling Disorder                            \\
GPT          & Generative Pretrained Transformer            \\
HDRS         & Hamilton Depression Rating Scale             \\
ICBT         & Internet-Based Cognitive Behavioral Therapy  \\
IECBT        & Internet-enabled CBT                         \\
LiCBT        & Low-Intensity Cognitive Behavioral Therapy   \\
LLM          & Large Language Model                         \\
MHA          & Mental Health America                        \\
MUBS         & Mobile-based Behavioral Activation System    \\
NLP          & Natural Language Processing                  \\
OCD          & Obsessive-Compulsive Disorder                \\
PD           & Panic disorder                               \\
PDT          & Psychodynamic Therapy                        \\
PTM          & Pre-trained Language Model                   \\
PTSD         & Post-Traumatic Stress Disorder               \\
Q\&A         & Question and Answering                       \\
ReHo         & Regional Homogeneity                         \\
RET          & Rational Emotive Therapy                     \\
SAD          & Social Anxiety Disorder                      \\
THI scores   & Tinnitus Handicap Inventory scores           \\
VR           & Virtual Reality                              \\
VRET         & Virtual Reality Exposure Therapy             \\
XAI          & Explainable Artificial Intelligence          \\

\bottomrule
\end{tabular}
\end{table}

\end{appendices}

\bibliography{references.bib}

\begin{thebibliography}{172}
\providecommand{\natexlab}[1]{#1}
\providecommand{\url}[1]{{#1}}
\providecommand{\urlprefix}{URL }
\providecommand{\doi}[1]{\url{https://doi.org/#1}}
\providecommand{\eprint}[2][]{\url{#2}}
 \bibcommenthead

\bibitem[{Abd-Alrazaq et~al(2019)Abd-Alrazaq, Alajlani, Alalwan, Bewick, Gardner, and Househ}]{abd2019overview}
Abd-Alrazaq AA, Alajlani M, Alalwan AA, et~al (2019) An overview of the features of chatbots in mental health: A scoping review. International journal of medical informatics 132:103978

\bibitem[{Abramowitz(2013)}]{abramowitz2013practice}
Abramowitz JS (2013) The practice of exposure therapy: Relevance of cognitive-behavioral theory and extinction theory. Behavior therapy 44(4):548--558

\bibitem[{Addepally and Purkayastha(2017)}]{addepally2017mobile}
Addepally SA, Purkayastha S (2017) Mobile-application based cognitive behavior therapy (cbt) for identifying and managing depression and anxiety. In: Digital Human Modeling. Applications in Health, Safety, Ergonomics, and Risk Management: Health and Safety: 8th International Conference, DHM 2017, Held as Part of HCI International 2017, Vancouver, BC, Canada, July 9-14, 2017, Proceedings, Part II 8, Springer, pp 3--12

\bibitem[{{\AA}hs et~al(2020){\AA}hs, Mozelius, and Dobslaw}]{aahs2020artificial}
{\AA}hs F, Mozelius P, Dobslaw F (2020) Artificial intelligence supported cognitive behavioral therapy for treatment of speech anxiety in virtual reality environments. In: ECIAIR 2020, Academic Conferences and Publishing International Limited

\bibitem[{Alhaj et~al(2022)Alhaj, Al-Haj, Sharieh, and Jabri}]{alhaj2022improving}
Alhaj F, Al-Haj A, Sharieh A, et~al (2022) Improving arabic cognitive distortion classification in twitter using bertopic. International Journal of Advanced Computer Science and Applications 13(1):854--860

\bibitem[{Alsharif and Philip(2015)}]{alsharif2015cognitive}
Alsharif AH, Philip N (2015) Cognitive behavioural therapy embedding smoking cessation program using smart phone technologies. In: 2015 5th World Congress on Information and Communication Technologies (WICT), IEEE, pp 134--139

\bibitem[{Ardulov et~al(2022)Ardulov, Creed, Atkins, and Narayanan}]{ardulov2022local}
Ardulov V, Creed TA, Atkins DC, et~al (2022) Local dynamic mode of cognitive behavioral therapy. arXiv preprint arXiv:220509752

\bibitem[{Assun{\c{c}}{\~a}o et~al(2022)Assun{\c{c}}{\~a}o, Patr{\~a}o, Castelo-Branco, and Menezes}]{assunccao2022overview}
Assun{\c{c}}{\~a}o G, Patr{\~a}o B, Castelo-Branco M, et~al (2022) An overview of emotion in artificial intelligence. IEEE Transactions on Artificial Intelligence 3(6):867--886

\bibitem[{{Australian National University}(2024)}]{MoodGym}
{Australian National University} (2024) Welcome to moodgym. https://www.moodgym.com.au/, accessed: 2024-06-19

\bibitem[{Ball et~al(2014)Ball, Stein, Ramsawh, Campbell-Sills, and Paulus}]{ball2014single}
Ball TM, Stein MB, Ramsawh HJ, et~al (2014) Single-subject anxiety treatment outcome prediction using functional neuroimaging. Neuropsychopharmacology 39(5):1254--1261

\bibitem[{Bandelow et~al(2017)Bandelow, Michaelis, and Wedekind}]{bandelow2017treatment}
Bandelow B, Michaelis S, Wedekind D (2017) Treatment of anxiety disorders. Dialogues in clinical neuroscience 19(2):93--107

\bibitem[{Bech et~al(1996)Bech, Steer, and Brown}]{bech1996beck}
Bech A, Steer R, Brown G (1996) Beck depression inventory—second edition: manual. San Antonio: The Psychological Corporation 4:561--71

\bibitem[{Beck(1979)}]{beck1979cognitive}
Beck AT (1979) Cognitive therapy and the emotional disorders. Penguin

\bibitem[{Beck et~al(1988)Beck, Epstein, Brown, and Steer}]{beck1988inventory}
Beck AT, Epstein N, Brown G, et~al (1988) An inventory for measuring clinical anxiety: psychometric properties. Journal of consulting and clinical psychology 56(6):893

\bibitem[{Beck and Beck(2011)}]{beck2011cognitive}
Beck JS, Beck AT (2011) Cognitive behavior therapy. New York: Basics and beyond Guilford Publication pp 19--20

\bibitem[{Benard and Lukandu(2018)}]{benard2018q}
Benard AO, Lukandu IA (2018) A q-learning model for cognitive behavioural therapy of insomnia patients. Int J Comput Inf Technol 7(3):1--7

\bibitem[{Bhaumik et~al(2023)Bhaumik, Srivastava, Jalali, Ghosh, and Chandrasekaran}]{bhaumik2023mindwatch}
Bhaumik R, Srivastava V, Jalali A, et~al (2023) Mindwatch: A smart cloud-based ai solution for suicide ideation detection leveraging large language models. medRxiv pp 2023--09

\bibitem[{Brave et~al(2005)Brave, Nass, and Hutchinson}]{brave2005computers}
Brave S, Nass C, Hutchinson K (2005) Computers that care: investigating the effects of orientation of emotion exhibited by an embodied computer agent. International journal of human-computer studies 62(2):161--178

\bibitem[{Burns and Beck(1999)}]{burns1999feeling}
Burns DD, Beck AT (1999) Feeling good: The new mood therapy. Avon New York

\bibitem[{Burns and Spangler(2000)}]{burns2000does}
Burns DD, Spangler DL (2000) Does psychotherapy homework lead to improvements in depression in cognitive--behavioral therapy or does improvement lead to increased homework compliance? Journal of consulting and clinical psychology 68(1):46

\bibitem[{Cardi et~al(2013)Cardi, Clarke, and Treasure}]{cardi2013use}
Cardi V, Clarke A, Treasure J (2013) The use of guided self-help incorporating a mobile component in people with eating disorders: A pilot study. European Eating Disorders Review 21(4):315--322

\bibitem[{Chandra et~al(2023)Chandra, Joshi, and Bhagwat}]{chandra2023chatgpt}
Chandra P, Joshi G, Bhagwat R (2023) Chatgpt's evolution in reshaping cognitive behavioral therapy. In: 2023 IEEE Engineering Informatics, IEEE, pp 1--9

\bibitem[{Chen et~al(2022{\natexlab{a}})Chen, Flemotomos, Imel, Atkins, and Narayanan}]{chen2022leveraging}
Chen Z, Flemotomos N, Imel ZE, et~al (2022{\natexlab{a}}) Leveraging open data and task augmentation to automated behavioral coding of psychotherapy conversations in low-resource scenarios. arXiv preprint arXiv:221014254

\bibitem[{Chen et~al(2022{\natexlab{b}})Chen, Flemotomos, Singla, Creed, Atkins, and Narayanan}]{chen2022automated}
Chen Z, Flemotomos N, Singla K, et~al (2022{\natexlab{b}}) An automated quality evaluation framework of psychotherapy conversations with local quality estimates. Computer speech \& language 75:101380

\bibitem[{Chen et~al(2023)Chen, Lu, and Wang}]{chen2023empowering}
Chen Z, Lu Y, Wang WY (2023) Empowering psychotherapy with large language models: Cognitive distortion detection through diagnosis of thought prompting. arXiv preprint arXiv:231007146

\bibitem[{{China Cognitive Behavioral therapy professional organization}(2024)}]{30DaySelfService}
{China Cognitive Behavioral therapy professional organization} (2024) China's first ccbt psychological self-service platform. https://www.psy.com.cn/therapy/, accessed: 2024-06-19

\bibitem[{Christ et~al(2020)Christ, Schouten, Blankers, van Schaik, Beekman, Wisman, Stikkelbroek, and Dekker}]{christ2020internet}
Christ C, Schouten MJ, Blankers M, et~al (2020) Internet and computer-based cognitive behavioral therapy for anxiety and depression in adolescents and young adults: systematic review and meta-analysis. Journal of medical Internet research 22(9):e17831

\bibitem[{Clark(2013)}]{clark2013cognitive}
Clark DA (2013) Cognitive restructuring. The Wiley handbook of cognitive behavioral therapy pp 1--22

\bibitem[{C{\^o}t{\'e}-Allard et~al(2022)C{\^o}t{\'e}-Allard, Pham, Schultz, Nordgreen, and Torresen}]{cote2022adherence}
C{\^o}t{\'e}-Allard U, Pham MH, Schultz AK, et~al (2022) Adherence forecasting for guided internet-delivered cognitive behavioral therapy: A minimally data-sensitive approach. IEEE Journal of Biomedical and Health Informatics

\bibitem[{Cuijpers(1998)}]{cuijpers1998psychoeducational}
Cuijpers P (1998) A psychoeducational approach to the treatment of depression: a meta-analysis of lewinsohn's “coping with depression” course. Behavior therapy 29(3):521--533

\bibitem[{Dahne et~al(2019)Dahne, Collado, Lejuez, Risco, Diaz, Coles, Kustanowitz, Zvolensky, and Carpenter}]{dahne2019pilot}
Dahne J, Collado A, Lejuez C, et~al (2019) Pilot randomized controlled trial of a spanish-language behavioral activation mobile app (!` apt{\'\i}vate!) for the treatment of depressive symptoms among united states latinx adults with limited english proficiency. Journal of Affective Disorders 250:210--217

\bibitem[{David et~al(2018)David, Cristea, and Hofmann}]{david2018cognitive}
David D, Cristea I, Hofmann SG (2018) Why cognitive behavioral therapy is the current gold standard of psychotherapy. Frontiers in psychiatry 9:333730

\bibitem[{Delgadillo and Gonzalez Salas~Duhne(2020)}]{delgadillo2020targeted}
Delgadillo J, Gonzalez Salas~Duhne P (2020) Targeted prescription of cognitive--behavioral therapy versus person-centered counseling for depression using a machine learning approach. Journal of Consulting and Clinical Psychology 88(1):14

\bibitem[{Demszky et~al(2023{\natexlab{a}})Demszky, Yang, Yeager, Bryan, Clapper, Chandhok, Eichstaedt, Hecht, Jamieson, and Johnson}]{2023Using}
Demszky D, Yang D, Yeager DS, et~al (2023{\natexlab{a}}) Using large language models in psychology. Nature Reviews Psychology 2(11):688--701

\bibitem[{Demszky et~al(2023{\natexlab{b}})Demszky, Yang, Yeager, Bryan, Clapper, Chandhok, Eichstaedt, Hecht, Jamieson, Johnson et~al}]{demszky2023using}
Demszky D, Yang D, Yeager DS, et~al (2023{\natexlab{b}}) Using large language models in psychology. Nature Reviews Psychology 2(11):688--701

\bibitem[{Denecke et~al(2022)Denecke, Schmid, N{\"u}ssli et~al}]{denecke2022implementation}
Denecke K, Schmid N, N{\"u}ssli S, et~al (2022) Implementation of cognitive behavioral therapy in e--mental health apps: Literature review. Journal of medical Internet research 24(3):e27791

\bibitem[{DiMatteo et~al(2002)DiMatteo, Giordani, Lepper, and Croghan}]{dimatteo2002patient}
DiMatteo MR, Giordani PJ, Lepper HS, et~al (2002) Patient adherence and medical treatment outcomes: a meta-analysis. Medical care 40(9):794--811

\bibitem[{Dimidjian et~al(2006)Dimidjian, Hollon, Dobson, Schmaling, Kohlenberg, Addis, Gallop, McGlinchey, Markley, Gollan et~al}]{dimidjian2006randomized}
Dimidjian S, Hollon SD, Dobson KS, et~al (2006) Randomized trial of behavioral activation, cognitive therapy, and antidepressant medication in the acute treatment of adults with major depression. Journal of consulting and clinical psychology 74(4):658

\bibitem[{Ding et~al(2022)Ding, Lybarger, Tauscher, and Cohen}]{ding2022improving}
Ding X, Lybarger K, Tauscher J, et~al (2022) Improving classification of infrequent cognitive distortions: domain-specific model vs. data augmentation. In: Proceedings of the 2022 conference of the North American chapter of the association for computational linguistics: human language technologies: Student Research Workshop, pp 68--75

\bibitem[{Dobson and Dozois(2021)}]{dobson2021handbook}
Dobson KS, Dozois DJ (2021) Handbook of cognitive-behavioral therapies. Guilford Publications

\bibitem[{Driessen and Hollon(2010)}]{driessen2010cognitive}
Driessen E, Hollon SD (2010) Cognitive behavioral therapy for mood disorders: efficacy, moderators and mediators. Psychiatric Clinics 33(3):537--555

\bibitem[{Elsharawi and El~Bolock(2024)}]{elsharawi2024c}
Elsharawi N, El~Bolock A (2024) C-journal: A journaling application for detecting and classifying cognitive distortions using deep-learning based on a crowd-sourced dataset. In: Proceedings of the 2024 Joint International Conference on Computational Linguistics, Language Resources and Evaluation (LREC-COLING 2024), pp 3224--3234

\bibitem[{Elyoseph et~al(2023)Elyoseph, Hadar-Shoval, Asraf, and Lvovsky}]{elyoseph2023chatgpt}
Elyoseph Z, Hadar-Shoval D, Asraf K, et~al (2023) Chatgpt outperforms humans in emotional awareness evaluations. Frontiers in Psychology 14:1199058

\bibitem[{Ewbank et~al(2020)Ewbank, Cummins, Tablan, Bateup, Catarino, Martin, and Blackwell}]{ewbank2020quantifying}
Ewbank MP, Cummins R, Tablan V, et~al (2020) Quantifying the association between psychotherapy content and clinical outcomes using deep learning. JAMA psychiatry 77(1):35--43

\bibitem[{Fitzpatrick et~al(2017)Fitzpatrick, Darcy, and Vierhile}]{fitzpatrick2017delivering}
Fitzpatrick KK, Darcy A, Vierhile M (2017) Delivering cognitive behavior therapy to young adults with symptoms of depression and anxiety using a fully automated conversational agent (woebot): a randomized controlled trial. JMIR mental health 4(2):e7785

\bibitem[{Flemotomos et~al(2021)Flemotomos, Martinez, Chen, Creed, Atkins, and Narayanan}]{flemotomos2021automated}
Flemotomos N, Martinez VR, Chen Z, et~al (2021) Automated quality assessment of cognitive behavioral therapy sessions through highly contextualized language representations. PloS one 16(10):e0258639

\bibitem[{Flemotomos et~al(2022)Flemotomos, Martinez, Chen, Singla, Ardulov, Peri, Caperton, Gibson, Tanana, Georgiou et~al}]{flemotomos2022automated}
Flemotomos N, Martinez VR, Chen Z, et~al (2022) Automated evaluation of psychotherapy skills using speech and language technologies. Behavior Research Methods 54(2):690--711

\bibitem[{Foa et~al(1993)Foa, Riggs, Dancu, and Rothbaum}]{foa1993reliability}
Foa EB, Riggs DS, Dancu CV, et~al (1993) Reliability and validity of a brief instrument for assessing post-traumatic stress disorder. Journal of traumatic stress 6(4):459--473

\bibitem[{Foreman and Pollard(2016)}]{foreman2016cognitive}
Foreman EI, Pollard C (2016) Cognitive Behavioural Therapy (CBT): Your Toolkit to Modify Mood, Overcome Obstructions and Improve Your Life. Icon Books, Limited

\bibitem[{Fu et~al(2023)Fu, Zhao, Li, Luo, Song, Zhai, Liu, Wang, Wang, Cheng et~al}]{fu2023enhancing}
Fu G, Zhao Q, Li J, et~al (2023) Enhancing psychological counseling with large language model: A multifaceted decision-support system for non-professionals. arXiv preprint arXiv:230815192

\bibitem[{Furukawa et~al(2023)Furukawa, Iwata, Horikoshi, Sakata, Toyomoto, Luo, Tajika, Kudo, and Aramaki}]{furukawa2023harnessing}
Furukawa TA, Iwata S, Horikoshi M, et~al (2023) Harnessing ai to optimize thought records and facilitate cognitive restructuring in smartphone cbt: An exploratory study. Cognitive Therapy and Research 47(6):887--893

\bibitem[{Garcia-Ceja et~al(2015)Garcia-Ceja, Osmani, and Mayora}]{garcia2015automatic}
Garcia-Ceja E, Osmani V, Mayora O (2015) Automatic stress detection in working environments from smartphones’ accelerometer data: a first step. IEEE journal of biomedical and health informatics 20(4):1053--1060

\bibitem[{Giordano et~al(2022)Giordano, Donati, Zamboni, Fusina, Primi, and Lugoboni}]{giordano2022alter}
Giordano R, Donati MA, Zamboni L, et~al (2022) Alter game: A study protocol on a virtual “serious game” for relapse prevention in patients with gambling disorder. Frontiers in Psychiatry 13:854088

\bibitem[{Goodwin et~al(2019)Goodwin, Mazefsky, Ioannidis, Erdogmus, and Siegel}]{goodwin2019predicting}
Goodwin MS, Mazefsky CA, Ioannidis S, et~al (2019) Predicting aggression to others in youth with autism using a wearable biosensor. Autism research 12(8):1286--1296

\bibitem[{Graham et~al(2019)Graham, Depp, Lee, Nebeker, Tu, Kim, and Jeste}]{graham2019artificial}
Graham S, Depp C, Lee EE, et~al (2019) Artificial intelligence for mental health and mental illnesses: an overview. Current psychiatry reports 21:1--18

\bibitem[{Gunlicks-Stoessel et~al(2020)Gunlicks-Stoessel, Klimes-Dougan, VanZomeren, and Ma}]{gunlicks2020developing}
Gunlicks-Stoessel M, Klimes-Dougan B, VanZomeren A, et~al (2020) Developing a data-driven algorithm for guiding selection between cognitive behavioral therapy, fluoxetine, and combination treatment for adolescent depression. Translational psychiatry 10(1):321

\bibitem[{Hahn et~al(2015)Hahn, Kircher, Straube, Wittchen, Konrad, Str{\"o}hle, Wittmann, Pfleiderer, Reif, Arolt et~al}]{hahn2015predicting}
Hahn T, Kircher T, Straube B, et~al (2015) Predicting treatment response to cognitive behavioral therapy in panic disorder with agoraphobia by integrating local neural information. JAMA psychiatry 72(1):68--74

\bibitem[{Harrison et~al(2011)Harrison, Proudfoot, Wee, Parker, Pavlovic, and Manicavasagar}]{harrison2011mobile}
Harrison V, Proudfoot J, Wee PP, et~al (2011) Mobile mental health: review of the emerging field and proof of concept study. Journal of mental health 20(6):509--524

\bibitem[{Heapy et~al(2017)Heapy, Higgins, Goulet, LaChappelle, Driscoll, Czlapinski, Buta, Piette, Krein, and Kerns}]{heapy2017interactive}
Heapy AA, Higgins DM, Goulet JL, et~al (2017) Interactive voice response--based self-management for chronic back pain: the copes noninferiority randomized trial. JAMA Internal Medicine 177(6):765--773

\bibitem[{Helbig and Fehm(2004)}]{helbig2004problems}
Helbig S, Fehm L (2004) Problems with homework in cbt: Rare exception or rather frequent? Behavioural and cognitive psychotherapy 32(3):291--301

\bibitem[{Heng(2021)}]{heng2021rewind}
Heng YK (2021) Rewind: psychoeducation game leveraging cognitive behavioral therapy (cbt) to enhance emotion control for generalized anxiety disorder. In: Extended Abstracts of the 2021 CHI Conference on Human Factors in Computing Systems, pp 1--5

\bibitem[{Hilbert et~al(2020)Hilbert, Kunas, Lueken, Kathmann, Fydrich, and Fehm}]{hilbert2020predicting}
Hilbert K, Kunas SL, Lueken U, et~al (2020) Predicting cognitive behavioral therapy outcome in the outpatient sector based on clinical routine data: A machine learning approach. Behaviour research and therapy 124:103530

\bibitem[{Hilbert et~al(2021)Hilbert, Jacobi, Kunas, Elsner, Reuter, Lueken, and Kathmann}]{hilbert2021identifying}
Hilbert K, Jacobi T, Kunas SL, et~al (2021) Identifying cbt non-response among ocd outpatients: A machine-learning approach. Psychotherapy Research 31(1):52--62

\bibitem[{Huguet et~al(2016)Huguet, Rao, McGrath, Wozney, Wheaton, Conrod, and Rozario}]{huguet2016systematic}
Huguet A, Rao S, McGrath PJ, et~al (2016) A systematic review of cognitive behavioral therapy and behavioral activation apps for depression. PloS one 11(5):e0154248

\bibitem[{Isacsson et~al(2023)Isacsson, Abdesslem, Forsell, Boman, and Kaldo}]{isacsson2023machine}
Isacsson NH, Abdesslem FB, Forsell E, et~al (2023) Machine learning predictions of outcome in internet-based cognitive behavioral therapy: methodological choices and clinical usefulness. Preprint posted online on April

\bibitem[{Izumi et~al(2024)Izumi, Tanaka, Shidara, Adachi, Kanayama, Kudo, and Nakamura}]{izumi2024response}
Izumi K, Tanaka H, Shidara K, et~al (2024) Response generation for cognitive behavioral therapy with large language models: Comparative study with socratic questioning. arXiv preprint arXiv:240115966

\bibitem[{Jacobson et~al(1996)Jacobson, Dobson, Truax, Addis, Koerner, Gollan, Gortner, and Prince}]{jacobson1996component}
Jacobson NS, Dobson KS, Truax PA, et~al (1996) A component analysis of cognitive-behavioral treatment for depression. Journal of Consulting and Clinical Psychology 64(2):295

\bibitem[{Jameel(2020)}]{jameel2020virtual}
Jameel L (2020) Virtual-reality assisted cbt for social difficulties: a feasibility study in early intervention for psychosis services. PhD thesis, King's College London

\bibitem[{Jang et~al(2021)Jang, Kim, Kim, Hong, Kim, and Kim}]{jang2021mobile}
Jang S, Kim JJ, Kim SJ, et~al (2021) Mobile app-based chatbot to deliver cognitive behavioral therapy and psychoeducation for adults with attention deficit: A development and feasibility/usability study. International journal of medical informatics 150:104440

\bibitem[{Jeong et~al(2024)Jeong, Song, Yang, and Kang}]{jeong2024advancing}
Jeong Y, Song JJ, Yang J, et~al (2024) Advancing tinnitus therapeutics: Gpt-2 driven clustering analysis of cognitive behavioral therapy sessions and google t5-based predictive modeling for thi score assessment. IEEE Access

\bibitem[{Jiang et~al(2024)Jiang, Yu, Zhao, Li, Song, Qi, Zhai, Luo, Wang, Fu et~al}]{jiang2024ai}
Jiang M, Yu YJ, Zhao Q, et~al (2024) Ai-enhanced cognitive behavioral therapy: Deep learning and large language models for extracting cognitive pathways from social media texts. arXiv preprint arXiv:240411449

\bibitem[{Kaldo et~al(2021)Kaldo, Isacsson, Forsell, Bjurner, Abdesslem, and Boman}]{kaldo2021ai}
Kaldo V, Isacsson N, Forsell E, et~al (2021) Ai-driven adaptive treatment strategies in internet-delivered cbt. European Psychiatry 64(S1):S20--S20

\bibitem[{Kazantzis et~al(2010)Kazantzis, Arntz, Borkovec, Holmes, and Wade}]{kazantzis2010unresolved}
Kazantzis N, Arntz AR, Borkovec T, et~al (2010) Unresolved issues regarding homework assignments in cognitive and behavioural therapies: An expert panel discussion at aacbt. Behaviour Change 27(3):119--129

\bibitem[{Khare et~al(2023)Khare, Blanes-Vidal, Nadimi, and Acharya}]{khare2023emotion}
Khare SK, Blanes-Vidal V, Nadimi ES, et~al (2023) Emotion recognition and artificial intelligence: A systematic review (2014--2023) and research recommendations. Information Fusion p 102019

\bibitem[{Koz{\l}owski et~al(2023)Koz{\l}owski, Gabor-Siatkowska, Stefaniak, Sowa{\'n}ski, and Janicki}]{kozlowski2023enhanced}
Koz{\l}owski M, Gabor-Siatkowska K, Stefaniak I, et~al (2023) Enhanced emotion and sentiment recognition for empathetic dialogue system using big data and deep learning methods. In: International Conference on Computational Science, Springer, pp 465--480

\bibitem[{Lan et~al(2019)Lan, Chen, Goodman, Gimpel, Sharma, and Soricut}]{lan2019albert}
Lan Z, Chen M, Goodman S, et~al (2019) Albert: A lite bert for self-supervised learning of language representations. arXiv preprint arXiv:190911942

\bibitem[{LeBeau et~al(2013)LeBeau, Davies, Culver, and Craske}]{lebeau2013homework}
LeBeau RT, Davies CD, Culver NC, et~al (2013) Homework compliance counts in cognitive-behavioral therapy. Cognitive behaviour therapy 42(3):171--179

\bibitem[{Lee et~al(2021)Lee, Torous, De~Choudhury, Depp, Graham, Kim, Paulus, Krystal, and Jeste}]{lee2021artificial}
Lee EE, Torous J, De~Choudhury M, et~al (2021) Artificial intelligence for mental health care: clinical applications, barriers, facilitators, and artificial wisdom. Biological Psychiatry: Cognitive Neuroscience and Neuroimaging 6(9):856--864

\bibitem[{Lee et~al(2024)Lee, Kang, Kim, Chung, Lee, and Yeo}]{lee2024cocoa}
Lee S, Kang J, Kim H, et~al (2024) Cocoa: Cbt-based conversational counseling agent using memory specialized in cognitive distortions and dynamic prompt. arXiv preprint arXiv:240217546

\bibitem[{Lee et~al(2023)Lee, Lee, Shin, Bae, and Hahn}]{lee2023chain}
Lee YK, Lee I, Shin M, et~al (2023) Chain of empathy: Enhancing empathetic response of large language models based on psychotherapy models. arXiv preprint arXiv:231104915

\bibitem[{Lewinsohn(1974)}]{lewinsohn1974behavioral}
Lewinsohn PM (1974) A behavioral approach to depression.

\bibitem[{Li et~al(2024)Li, Herderich, and Goldenberg}]{li2024skill}
Li JZ, Herderich A, Goldenberg A (2024) Skill but not effort drive gpt overperformance over humans in cognitive reframing of negative scenarios. Preprint posted online on April

\bibitem[{Lim et~al(2024)Lim, Kim, Choi, Sohn, and Kim}]{lim2024erd}
Lim S, Kim Y, Choi CH, et~al (2024) Erd: A framework for improving llm reasoning for cognitive distortion classification. arXiv preprint arXiv:240314255

\bibitem[{Lin et~al(2024)Lin, Wang, Dong, and Ni}]{lin2024detection}
Lin S, Wang Y, Dong J, et~al (2024) Detection and positive reconstruction of cognitive distortion sentences: Mandarin dataset and evaluation. arXiv preprint arXiv:240515334

\bibitem[{Linardon et~al(2017)Linardon, Wade, De~la Piedad~Garcia, and Brennan}]{linardon2017efficacy}
Linardon J, Wade TD, De~la Piedad~Garcia X, et~al (2017) The efficacy of cognitive-behavioral therapy for eating disorders: A systematic review and meta-analysis. Journal of consulting and clinical psychology 85(11):1080

\bibitem[{Lorimer et~al(2021)Lorimer, Delgadillo, Kellett, and Lawrence}]{lorimer2021dynamic}
Lorimer B, Delgadillo J, Kellett S, et~al (2021) Dynamic prediction and identification of cases at risk of relapse following completion of low-intensity cognitive behavioural therapy. Psychotherapy Research 31(1):19--32

\bibitem[{Maddela et~al(2023)Maddela, Ung, Xu, Madotto, Foran, and Boureau}]{maddela2023training}
Maddela M, Ung M, Xu J, et~al (2023) Training models to generate, recognize, and reframe unhelpful thoughts. In: Proceedings of the 61st Annual Meeting of the Association for Computational Linguistics (Volume 1: Long Papers), pp 13641--13660

\bibitem[{Madhu et~al(2022)Madhu, Kumar, Pal, and Rubini}]{madhu2022activity}
Madhu SH, Kumar SS, Pal M, et~al (2022) Activity recognition for behavioral activation in depression with artificial intelligence. In: 2022 IEEE 4th PhD Colloquium on Emerging Domain Innovation and Technology for Society (PhD EDITS), IEEE, pp 1--2

\bibitem[{Malgaroli et~al(2023)Malgaroli, Hull, Zech, and Althoff}]{malgaroli2023natural}
Malgaroli M, Hull TD, Zech JM, et~al (2023) Natural language processing for mental health interventions: a systematic review and research framework. Translational Psychiatry 13(1):309

\bibitem[{M{\aa}nsson et~al(2015)M{\aa}nsson, Frick, Boraxbekk, Marquand, Williams, Carlbring, Andersson, and Furmark}]{maansson2015predicting}
M{\aa}nsson KN, Frick A, Boraxbekk CJ, et~al (2015) Predicting long-term outcome of internet-delivered cognitive behavior therapy for social anxiety disorder using fmri and support vector machine learning. Translational psychiatry 5(3):e530--e530

\bibitem[{Maples-Keller et~al(2017)Maples-Keller, Bunnell, Kim, and Rothbaum}]{maples2017use}
Maples-Keller JL, Bunnell BE, Kim SJ, et~al (2017) The use of virtual reality technology in the treatment of anxiety and other psychiatric disorders. Harvard review of psychiatry 25(3):103--113

\bibitem[{Mart{\'\i}nez-Miranda et~al(2019)Mart{\'\i}nez-Miranda, Mart{\'\i}nez, Ramos, Aguilar, Jim{\'e}nez, Arias, Rosales, and Valencia}]{martinez2019assessment}
Mart{\'\i}nez-Miranda J, Mart{\'\i}nez A, Ramos R, et~al (2019) Assessment of users’ acceptability of a mobile-based embodied conversational agent for the prevention and detection of suicidal behaviour. Journal of medical systems 43(8):246

\bibitem[{McGinn and Sanderson(2001)}]{mcginn2001allows}
McGinn LK, Sanderson WC (2001) What allows cognitive behavioral therapy to be brief: Overview, efficacy, and crucial factors facilitating brief treatment. Clinical psychology: science and practice 8(1):23

\bibitem[{McHugh et~al(2010)McHugh, Hearon, and Otto}]{mchugh2010cognitive}
McHugh RK, Hearon BA, Otto MW (2010) Cognitive behavioral therapy for substance use disorders. Psychiatric Clinics 33(3):511--525

\bibitem[{Mehta et~al(2021)Mehta, Niles, Vargas, Marafon, Couto, and Gross}]{mehta2021acceptability}
Mehta A, Niles AN, Vargas JH, et~al (2021) Acceptability and effectiveness of artificial intelligence therapy for anxiety and depression (youper): Longitudinal observational study. Journal of medical Internet research 23(6):e26771

\bibitem[{Michelle et~al(2014)Michelle, Jarzabek, and Wadhwa}]{michelle2014cbt}
Michelle TQY, Jarzabek S, Wadhwa B (2014) Cbt assistant: Mhealth app for psychotherapy. In: 2014 IEEE Global Humanitarian Technology Conference-South Asia Satellite (GHTC-SAS), IEEE, pp 135--140

\bibitem[{Moody et~al(2017)Moody, Morfini, Cheng, Sheen, Tadayonnejad, Reggente, O'neill, and Feusner}]{moody2017mechanisms}
Moody T, Morfini F, Cheng G, et~al (2017) Mechanisms of cognitive-behavioral therapy for obsessive-compulsive disorder involve robust and extensive increases in brain network connectivity. Translational psychiatry 7(9):e1230--e1230

\bibitem[{Mostafa et~al(2021)Mostafa, El~Bolock, and Abdennadher}]{mostafa2021automatic}
Mostafa M, El~Bolock A, Abdennadher S (2021) Automatic detection and classification of cognitive distortions in journaling text. In: WEBIST, pp 444--452

\bibitem[{Na(2024)}]{na2024cbt}
Na H (2024) Cbt-llm: A chinese large language model for cognitive behavioral therapy-based mental health question answering. arXiv preprint arXiv:240316008

\bibitem[{Nazarova(2023)}]{nazarova2023application}
Nazarova D (2023) Application of artificial intelligence in mental healthcare: Generative pre-trained transformer 3 (gpt-3) and cognitive distortions. In: Proceedings of the Future Technologies Conference, Springer, pp 204--219

\bibitem[{Nepal et~al(2024)Nepal, Pillai, Campbell, Massachi, Choi, Xu, Kuc, Huckins, Holden, Depp et~al}]{nepal2024contextual}
Nepal S, Pillai A, Campbell W, et~al (2024) Contextual ai journaling: Integrating llm and time series behavioral sensing technology to promote self-reflection and well-being using the mindscape app. arXiv preprint arXiv:240400487

\bibitem[{Nwoye et~al(2024)Nwoye, Muslehat, Umeh, Okodeh, and Woo}]{nwoye2024schizobot}
Nwoye EO, Muslehat AA, Umeh C, et~al (2024) Schizobot: Delivering cognitive behavioural therapy for augmented management of schizophrenia. Digital Technologies Research and Applications 3(2):24--40

\bibitem[{Olatunji et~al(2010)Olatunji, Cisler, and Deacon}]{olatunji2010efficacy}
Olatunji BO, Cisler JM, Deacon BJ (2010) Efficacy of cognitive behavioral therapy for anxiety disorders: a review of meta-analytic findings. Psychiatric Clinics 33(3):557--577

\bibitem[{Oliveira et~al(2021)Oliveira, Matos, Junior, and Delabrida}]{oliveira2021initial}
Oliveira ALS, Matos LN, Junior MC, et~al (2021) An initial assessment of a chatbot for rumination-focused cognitive behavioral therapy (rfcbt) in college students. In: Computational Science and Its Applications--ICCSA 2021: 21st International Conference, Cagliari, Italy, September 13--16, 2021, Proceedings, Part VI 21, Springer, pp 549--564

\bibitem[{{\O}rskov et~al(2022){\O}rskov, Lichtenstein, Ernst, Fasterholdt, Matthiesen, Scirea, Bouchard, and Andersen}]{orskov2022cognitive}
{\O}rskov PT, Lichtenstein MB, Ernst MT, et~al (2022) Cognitive behavioral therapy with adaptive virtual reality exposure vs. cognitive behavioral therapy with in vivo exposure in the treatment of social anxiety disorder: A study protocol for a randomized controlled trial. Frontiers in psychiatry 13:991755

\bibitem[{Ozomaro et~al(2013)Ozomaro, Wahlestedt, and Nemeroff}]{ozomaro2013personalized}
Ozomaro U, Wahlestedt C, Nemeroff CB (2013) Personalized medicine in psychiatry: problems and promises. BMC medicine 11:1--35

\bibitem[{Pan et~al(2019)Pan, Huang, Zhao, Wang, Wang, and Qian}]{pan2019comparison}
Pan MR, Huang F, Zhao MJ, et~al (2019) A comparison of efficacy between cognitive behavioral therapy (cbt) and cbt combined with medication in adults with attention-deficit/hyperactivity disorder (adhd). Psychiatry research 279:23--33

\bibitem[{Parry et~al(2012)Parry, Cooper, Moore, Yadegarfar, Campbell, Esmonde, Morice, and Hutchcroft}]{parry2012cognitive}
Parry GD, Cooper CL, Moore JM, et~al (2012) Cognitive behavioural intervention for adults with anxiety complications of asthma: prospective randomised trial. Respiratory Medicine 106(6):802--810

\bibitem[{Patel et~al(2019)Patel, Thakore, Nandwani, and Bharti}]{patel2019combating}
Patel F, Thakore R, Nandwani I, et~al (2019) Combating depression in students using an intelligent chatbot: a cognitive behavioral therapy. In: 2019 IEEE 16th India Council International Conference (INDICON), IEEE, pp 1--4

\bibitem[{Pei et~al(2022)Pei, He, Yang, Lv, Jiao, Meng, Yan, Cui, He, Zhou et~al}]{pei2022acupuncture}
Pei W, He T, Yang P, et~al (2022) Acupuncture combined with cognitive--behavioural therapy for insomnia (cbt-i) in patients with insomnia: study protocol for a randomised controlled trial. BMJ open 12(12):e063442

\bibitem[{Penninx et~al(2022)Penninx, Benros, Klein, and Vinkers}]{penninx2022covid}
Penninx BW, Benros ME, Klein RS, et~al (2022) How covid-19 shaped mental health: from infection to pandemic effects. Nature medicine 28(10):2027--2037

\bibitem[{Peretz et~al(2023)Peretz, Taylor, Ruzek, Jefroykin, and Sadeh-Sharvit}]{peretz2023machine}
Peretz G, Taylor CB, Ruzek JI, et~al (2023) Machine learning model to predict assignment of therapy homework in behavioral treatments: Algorithm development and validation. JMIR Formative Research 7:e45156

\bibitem[{Petersen et~al(2016)Petersen, Sprich, Wilhelm et~al}]{petersen2016massachusetts}
Petersen TJ, Sprich SE, Wilhelm S, et~al (2016) The massachusetts general hospital handbook of cognitive behavioral therapy. Tech. rep., Springer

\bibitem[{Piette et~al(2016)Piette, Krein, Striplin, Marinec, Kerns, Farris, Singh, An, and Heapy}]{piette2016patient}
Piette JD, Krein SL, Striplin D, et~al (2016) Patient-centered pain care using artificial intelligence and mobile health tools: protocol for a randomized study funded by the us department of veterans affairs health services research and development program. JMIR research protocols 5(2):e4995

\bibitem[{Prasad et~al(2023)Prasad, Chien, Regan, Enrique, Palacios, Keegan, Munir, Tanno, Richardson, Nori et~al}]{prasad2023deep}
Prasad N, Chien I, Regan T, et~al (2023) Deep learning for the prediction of clinical outcomes in internet-delivered cbt for depression and anxiety. Plos one 18(11):e0272685

\bibitem[{Provoost et~al(2019)Provoost, Ruwaard, Van~Breda, Riper, and Bosse}]{provoost2019validating}
Provoost S, Ruwaard J, Van~Breda W, et~al (2019) Validating automated sentiment analysis of online cognitive behavioral therapy patient texts: an exploratory study. Frontiers in psychology 10:432272

\bibitem[{Qi et~al(2023)Qi, Zhao, Song, Zhai, Luo, Liu, Yu, Wang, Zou, Yang et~al}]{qi2023evaluating}
Qi H, Zhao Q, Song C, et~al (2023) Evaluating the efficacy of supervised learning vs large language models for identifying cognitive distortions and suicidal risks in chinese social media. arXiv preprint arXiv:230903564

\bibitem[{Rahman et~al(2022)Rahman, Brown, Shopland, Harris, Turabee, Heym, Sumich, Standen, Downes, Xing et~al}]{rahman2022towards}
Rahman MA, Brown DJ, Shopland N, et~al (2022) Towards machine learning driven self-guided virtual reality exposure therapy based on arousal state detection from multimodal data. In: International Conference on Brain Informatics, Springer, pp 195--209

\bibitem[{Rajagopal et~al(2021)Rajagopal, Nirmala, Andrew, and Arun}]{rajagopal2021novel}
Rajagopal A, Nirmala V, Andrew J, et~al (2021) Novel ai to avert the mental health crisis in covid-19: Novel application of gpt2 in cognitive behaviour therapy. Preprint posted online on January

\bibitem[{Rani et~al(2023)Rani, Vishnoi, and Mishra}]{rani2023mental}
Rani K, Vishnoi H, Mishra M (2023) A mental health chatbot delivering cognitive behavior therapy and remote health monitoring using nlp and ai. In: 2023 International Conference on Disruptive Technologies (ICDT), IEEE, pp 313--317

\bibitem[{Rathje et~al(2023)Rathje, Mirea, Sucholutsky, Marjieh, Robertson, and Van~Bavel}]{rathje2023gpt}
Rathje S, Mirea D, Sucholutsky I, et~al (2023) Gpt is an effective tool for multilingual psychological text analysis. psyarxiv. Preprint posted online on July 17

\bibitem[{Rathnayaka et~al(2022)Rathnayaka, Mills, Burnett, De~Silva, Alahakoon, and Gray}]{rathnayaka2022mental}
Rathnayaka P, Mills N, Burnett D, et~al (2022) A mental health chatbot with cognitive skills for personalised behavioural activation and remote health monitoring. Sensors 22(10):3653

\bibitem[{Reger et~al(2013)Reger, Hoffman, Riggs, Rothbaum, Ruzek, Holloway, and Kuhn}]{reger2013pe}
Reger GM, Hoffman J, Riggs D, et~al (2013) The “pe coach” smartphone application: An innovative approach to improving implementation, fidelity, and homework adherence during prolonged exposure. Psychological services 10(3):342

\bibitem[{Reggente et~al(2018)Reggente, Moody, Morfini, Sheen, Rissman, O’Neill, and Feusner}]{reggente2018multivariate}
Reggente N, Moody TD, Morfini F, et~al (2018) Multivariate resting-state functional connectivity predicts response to cognitive behavioral therapy in obsessive--compulsive disorder. Proceedings of the National Academy of Sciences 115(9):2222--2227

\bibitem[{Rizea(2022)}]{rizea2022deep}
Rizea A (2022) Deep learning-based solution for mental health issues. DATABASE SYSTEMS p~57

\bibitem[{Rohani et~al(2020)Rohani, Quemada~Lopategui, Tuxen, Faurholt-Jepsen, Kessing, and Bardram}]{rohani2020mubs}
Rohani DA, Quemada~Lopategui A, Tuxen N, et~al (2020) Mubs: A personalized recommender system for behavioral activation in mental health. In: Proceedings of the 2020 CHI Conference on Human Factors in Computing Systems, pp 1--13

\bibitem[{Rose et~al(2013)Rose, Buckey~Jr, Zbozinek, Motivala, Glenn, Cartreine, and Craske}]{rose2013randomized}
Rose RD, Buckey~Jr JC, Zbozinek TD, et~al (2013) A randomized controlled trial of a self-guided, multimedia, stress management and resilience training program. Behaviour research and therapy 51(2):106--112

\bibitem[{Sabour et~al(2023)Sabour, Zhang, Xiao, Zhang, Zheng, Wen, Zhao, and Huang}]{sabour2023chatbot}
Sabour S, Zhang W, Xiao X, et~al (2023) A chatbot for mental health support: exploring the impact of emohaa on reducing mental distress in china. Frontiers in digital health 5:1133987

\bibitem[{Schabus et~al(2023)Schabus, Eigl, Topalidis, and Hinterberger}]{schabus2023efficacy}
Schabus M, Eigl ES, Topalidis P, et~al (2023) Efficacy of digital cognitive behavioural therapy for insomnia: A randomised controlled trial using a new app that tracks sleep continuously using hrv. In: World Sleep 2023

\bibitem[{Schwartz et~al(2021)Schwartz, Cohen, Rubel, Zimmermann, Wittmann, and Lutz}]{schwartz2021personalized}
Schwartz B, Cohen ZD, Rubel JA, et~al (2021) Personalized treatment selection in routine care: Integrating machine learning and statistical algorithms to recommend cognitive behavioral or psychodynamic therapy. Psychotherapy Research 31(1):33--51

\bibitem[{Scozzari and Gamberini(2011)}]{scozzari2011virtual}
Scozzari S, Gamberini L (2011) Virtual reality as a tool for cognitive behavioral therapy: a review. Advanced computational intelligence paradigms in healthcare 6 Virtual reality in psychotherapy, rehabilitation, and assessment pp 63--108

\bibitem[{Sharma et~al(2020)Sharma, Miner, Atkins, and Althoff}]{sharma2020computational}
Sharma A, Miner A, Atkins D, et~al (2020) A computational approach to understanding empathy expressed in text-based mental health support. In: Proceedings of the 2020 Conference on Empirical Methods in Natural Language Processing (EMNLP), pp 5263--5276

\bibitem[{Sharma et~al(2023{\natexlab{a}})Sharma, Lin, Miner, Atkins, and Althoff}]{sharma2023human}
Sharma A, Lin IW, Miner AS, et~al (2023{\natexlab{a}}) Human--ai collaboration enables more empathic conversations in text-based peer-to-peer mental health support. Nature Machine Intelligence 5(1):46--57

\bibitem[{Sharma et~al(2023{\natexlab{b}})Sharma, Rushton, Lin, Nguyen, and Althoff}]{sharma2023facilitating}
Sharma A, Rushton K, Lin IW, et~al (2023{\natexlab{b}}) Facilitating self-guided mental health interventions through human-language model interaction: A case study of cognitive restructuring. arXiv preprint arXiv:231015461

\bibitem[{Sharma et~al(2023{\natexlab{c}})Sharma, Rushton, Lin, Wadden, Lucas, Miner, Nguyen, and Althoff}]{sharma2023cognitive}
Sharma A, Rushton K, Lin IW, et~al (2023{\natexlab{c}}) Cognitive reframing of negative thoughts through human-language model interaction. arXiv preprint arXiv:230502466

\bibitem[{Shickel et~al(2020)Shickel, Siegel, Heesacker, Benton, and Rashidi}]{shickel2020automatic}
Shickel B, Siegel S, Heesacker M, et~al (2020) Automatic detection and classification of cognitive distortions in mental health text. In: 2020 IEEE 20th International Conference on Bioinformatics and Bioengineering (BIBE), IEEE, pp 275--280

\bibitem[{Shidara et~al(2022)Shidara, Tanaka, Adachi, Kanayama, Sakagami, Kudo, and Nakamura}]{shidara2022automatic}
Shidara K, Tanaka H, Adachi H, et~al (2022) Automatic thoughts and facial expressions in cognitive restructuring with virtual agents. Frontiers in Computer Science 4:762424

\bibitem[{Shreevastava and Foltz(2021)}]{shreevastava2021detecting}
Shreevastava S, Foltz P (2021) Detecting cognitive distortions from patient-therapist interactions. In: Proceedings of the Seventh Workshop on Computational Linguistics and Clinical Psychology: Improving Access, pp 151--158

\bibitem[{Shurick et~al(2012)Shurick, Hamilton, Harris, Roy, Gross, and Phelps}]{shurick2012durable}
Shurick AA, Hamilton JR, Harris LT, et~al (2012) Durable effects of cognitive restructuring on conditioned fear. Emotion 12(6):1393

\bibitem[{Siddiqua et~al(2023)Siddiqua, Islam, Bolaka, Khan, and Momen}]{siddiqua2023aida}
Siddiqua R, Islam N, Bolaka JF, et~al (2023) Aida: Artificial intelligence based depression assessment applied to bangladeshi students. Array 18:100291

\bibitem[{Simms et~al(2017)Simms, Ramstedt, Rich, Richards, Martinez, and Giraud-Carrier}]{simms2017detecting}
Simms T, Ramstedt C, Rich M, et~al (2017) Detecting cognitive distortions through machine learning text analytics. In: 2017 IEEE international conference on healthcare informatics (ICHI), IEEE, pp 508--512

\bibitem[{Singh et~al(2023)Singh, Ghosh, Ekbal, and Bhattacharyya}]{singh2023decode}
Singh GV, Ghosh S, Ekbal A, et~al (2023) Decode: Detection of cognitive distortion and emotion cause extraction in clinical conversations. In: European Conference on Information Retrieval, Springer, pp 156--171

\bibitem[{Stephenson et~al(2023)Stephenson, Jagayat, Kumar, Khamooshi, Eadie, Pannu, Meartsi, Danaee, Gutierrez, Khan et~al}]{stephenson2023comparing}
Stephenson C, Jagayat J, Kumar A, et~al (2023) Comparing clinical decision-making of ai technology to a multi-professional care team in an electronic cognitive behavioural therapy program for depression: protocol. Frontiers in Psychiatry 14:1220607

\bibitem[{Stirman et~al(2021)Stirman, Gutner, Gamarra, Suvak, Vogt, Johnson, Wachen, Dondanville, Yarvis, Mintz et~al}]{stirman2021novel}
Stirman SW, Gutner CA, Gamarra J, et~al (2021) A novel approach to the assessment of fidelity to a cognitive behavioral therapy for ptsd using clinical worksheets: A proof of concept with cognitive processing therapy. Behavior therapy 52(3):656--672

\bibitem[{Striegl et~al(2023)Striegl, Richter, Grossmann, Br{\aa}stad, Gotthardt, Ruck, Wallert, and Loitsch}]{striegl2023deep}
Striegl J, Richter JW, Grossmann L, et~al (2023) Deep learning-based dimensional emotion recognition for conversational agent-based cognitive behavioral therapy. Preprint posted online on December

\bibitem[{Su et~al(2022)Su, Wang, Jiang, Zhao, Gao, Wu, Tao, Su, Zhang, Li et~al}]{su2022efficacy}
Su S, Wang Y, Jiang W, et~al (2022) Efficacy of artificial intelligence-assisted psychotherapy in patients with anxiety disorders: a prospective, national multicenter randomized controlled trial protocol. Frontiers in Psychiatry 12:799917

\bibitem[{Sun et~al(2021)Sun, Lin, Zheng, Liu, and Huang}]{sun2021psyqa}
Sun H, Lin Z, Zheng C, et~al (2021) Psyqa: A chinese dataset for generating long counseling text for mental health support. In: Findings of the Association for Computational Linguistics: ACL-IJCNLP 2021, pp 1489--1503

\bibitem[{Tanana et~al(2021)Tanana, Soma, Kuo, Bertagnolli, Dembe, Pace, Srikumar, Atkins, and Imel}]{tanana2021you}
Tanana MJ, Soma CS, Kuo PB, et~al (2021) How do you feel? using natural language processing to automatically rate emotion in psychotherapy. Behavior research methods pp 1--14

\bibitem[{Tang et~al(2017)Tang, Kreindler et~al}]{tang2017supporting}
Tang W, Kreindler D, et~al (2017) Supporting homework compliance in cognitive behavioural therapy: essential features of mobile apps. JMIR mental health 4(2):e5283

\bibitem[{Tauscher et~al(2023)Tauscher, Lybarger, Ding, Chander, Hudenko, Cohen, and Ben-Zeev}]{tauscher2023automated}
Tauscher JS, Lybarger K, Ding X, et~al (2023) Automated detection of cognitive distortions in text exchanges between clinicians and people with serious mental illness. Psychiatric services 74(4):407--410

\bibitem[{de~Toledo~Rodriguez et~al(2021)de~Toledo~Rodriguez, Salton, and Ross}]{de2021formulating}
de~Toledo~Rodriguez I, Salton G, Ross R (2021) Formulating automated responses to cognitive distortions for cbt interactions. In: Proceedings of the 4th International Conference on Natural Language and Speech Processing (ICNLSP 2021), pp 108--116

\bibitem[{Tolmeijer et~al(2018)Tolmeijer, Kumari, Peters, Williams, and Mason}]{tolmeijer2018using}
Tolmeijer E, Kumari V, Peters E, et~al (2018) Using fmri and machine learning to predict symptom improvement following cognitive behavioural therapy for psychosis. NeuroImage: Clinical 20:1053--1061

\bibitem[{Turkington et~al(2004)Turkington, Dudley, Warman, and Beck}]{turkington2004cognitive}
Turkington D, Dudley R, Warman DM, et~al (2004) Cognitive-behavioral therapy for schizophrenia: a review. Journal of Psychiatric Practice{\textregistered} 10(1):5--16

\bibitem[{Tymofiyeva et~al(2019)Tymofiyeva, Yuan, Huang, Connolly, Blom, Xu, and Yang}]{tymofiyeva2019application}
Tymofiyeva O, Yuan JP, Huang CY, et~al (2019) Application of machine learning to structural connectome to predict symptom reduction in depressed adolescents with cognitive behavioral therapy (cbt). NeuroImage: Clinical 23:101914

\bibitem[{Vieira et~al(2022)Vieira, Liang, Guiomar, and Mechelli}]{vieira2022can}
Vieira S, Liang X, Guiomar R, et~al (2022) Can we predict who will benefit from cognitive-behavioural therapy? a systematic review and meta-analysis of machine learning studies. Clinical Psychology Review 97:102193

\bibitem[{Vuyyuru et~al(2023)Vuyyuru, Krishna, Mary, Kayalvili, and Alsubayhay}]{vuyyuru2023transformer}
Vuyyuru VA, Krishna GV, Mary SSC, et~al (2023) A transformer-cnn hybrid model for cognitive behavioral therapy in psychological assessment and intervention for enhanced diagnostic accuracy and treatment efficiency. International Journal of Advanced Computer Science and Applications 14(7)

\bibitem[{Wanderley~Espinola et~al(2022)Wanderley~Espinola, Gomes, M{\^o}nica Silva~Pereira, and dos Santos}]{wanderley2022detection}
Wanderley~Espinola C, Gomes JC, M{\^o}nica Silva~Pereira J, et~al (2022) Detection of major depressive disorder, bipolar disorder, schizophrenia and generalized anxiety disorder using vocal acoustic analysis and machine learning: an exploratory study. Research on Biomedical Engineering 38(3):813--829

\bibitem[{Wang et~al(2023{\natexlab{a}})Wang, Deng, Zhao, and Qin}]{wang2023c2d2}
Wang B, Deng P, Zhao Y, et~al (2023{\natexlab{a}}) C2d2 dataset: A resource for the cognitive distortion analysis and its impact on mental health. In: The 2023 Conference on Empirical Methods in Natural Language Processing

\bibitem[{Wang et~al(2023{\natexlab{b}})Wang, Zhao, Lu, and Qin}]{wang2023cognitive}
Wang B, Zhao Y, Lu X, et~al (2023{\natexlab{b}}) Cognitive distortion based explainable depression detection and analysis technologies for the adolescent internet users on social media. Frontiers in Public Health 10:1045777

\bibitem[{Wang et~al(2024{\natexlab{a}})Wang, Milani, Chiu, Eack, Labrum, Murphy, Jones, Hardy, Shen, Fang et~al}]{wang2024patient}
Wang R, Milani S, Chiu JC, et~al (2024{\natexlab{a}}) {PATIENT}-${\Psi}$: Using large language models to simulate patients for training mental health professionals. arXiv preprint arXiv:240519660

\bibitem[{Wang et~al(2024{\natexlab{b}})Wang, Sharma, and Kumar}]{wang2024cognitive}
Wang X, Sharma D, Kumar D (2024{\natexlab{b}}) Cognitive reframing via large language models for enhanced linguistic attributes. In: The Second Tiny Papers Track at ICLR 2024

\bibitem[{Webb et~al(2017)Webb, Rosso, and Rauch}]{webb2017internet}
Webb CA, Rosso IM, Rauch SL (2017) Internet-based cognitive-behavioral therapy for depression: current progress and future directions. Harvard review of psychiatry 25(3):114--122

\bibitem[{Wei et~al(2023)Wei, Zhang, Wang, Womer, Dong, Zheng, Zhang, and Wang}]{wei2023towards}
Wei Y, Zhang R, Wang Y, et~al (2023) Towards a neuroimaging biomarker for predicting cognitive behavioural therapy outcomes in treatment-naive depression: Preliminary findings. Psychiatry Research 329:115542

\bibitem[{Whiteside et~al(2014)Whiteside, Ale, Vickers~Douglas, Tiede, and Dammann}]{whiteside2014case}
Whiteside SP, Ale CM, Vickers~Douglas K, et~al (2014) Case examples of enhancing pediatric ocd treatment with a smartphone application. Clinical Case Studies 13(1):80--94

\bibitem[{Xiao et~al(2024)Xiao, Xie, Kuang, Liu, Yang, Peng, Han, and Huang}]{xiao2024healme}
Xiao M, Xie Q, Kuang Z, et~al (2024) Healme: Harnessing cognitive reframing in large language models for psychotherapy. arXiv preprint arXiv:240305574

\bibitem[{Xing et~al(2017)Xing, Zhao, and Miao}]{xing2017identifying}
Xing Z, Zhao X, Miao C (2017) Identifying cognitive distortion by convolutional neural network based text classification. International Journal of Information Technology

\bibitem[{Yang et~al(2018)Yang, Gao, Jiang, Jin, Gao, Ma, and Woo}]{yang2018iot}
Yang S, Gao B, Jiang L, et~al (2018) Iot structured long-term wearable social sensing for mental wellbeing. IEEE Internet of Things Journal 6(2):3652--3662

\bibitem[{Yin et~al(2024)Yin, Jia, and Wakslak}]{yin2024ai}
Yin Y, Jia N, Wakslak CJ (2024) Ai can help people feel heard, but an ai label diminishes this impact. Proceedings of the National Academy of Sciences 121(14):e2319112121

\bibitem[{Zhan et~al(2024)Zhan, Zheng, Lee, Suh, Li, and Ong}]{zhan2024large}
Zhan H, Zheng A, Lee YK, et~al (2024) Large language models are capable of offering cognitive reappraisal, if guided. arXiv preprint arXiv:240401288

\bibitem[{Zhang et~al(2018)Zhang, Dinan, Urbanek, Szlam, Kiela, and Weston}]{zhang2018personalizing}
Zhang S, Dinan E, Urbanek J, et~al (2018) Personalizing dialogue agents: I have a dog, do you have pets too? In: Proceedings of the 56th Annual Meeting of the Association for Computational Linguistics (Volume 1: Long Papers), pp 2204--2213

\bibitem[{Zhang et~al(2023)Zhang, Tanana, Weitzman, Narayanan, Atkins, and Imel}]{zhang2023you}
Zhang X, Tanana M, Weitzman L, et~al (2023) You never know what you are going to get: Large-scale assessment of therapists’ supportive counseling skill use. Psychotherapy 60(2):149

\bibitem[{Zheng et~al(2022)Zheng, Ye et~al}]{zheng2022prediction}
Zheng Y, Ye Y, et~al (2022) Prediction of cognitive-behavioral therapy using deep learning for the treatment of adolescent social anxiety and mental health conditions. Scientific Programming 2022

\end{thebibliography}
\end{document}